\documentclass[]{ceurart}

\usepackage{listings}
\usepackage{amsmath,amssymb}

\usepackage{subcaption}

\begin{document}

\newcommand{\geofila}[1]{\textcolor{blue!60!blue}{[GF: #1]}}

\copyrightyear{2023}
\copyrightclause{Copyright for this paper by its authors.
  Use permitted under Creative Commons License Attribution 4.0
  International (CC BY 4.0).}

\conference{In A. Martin, K. Hinkelmann, H.-G. Fill, A. Gerber, D. Lenat, R. Stolle, F. van Harmelen (Eds.), 
Proceedings of the AAAI 2023 Spring Symposium on Challenges Requiring the Combination of Machine Learning and Knowledge Engineering (AAAI-MAKE 2023), Hyatt Regency, San Francisco Airport, California, USA, March 27-29, 2023.}

\title{Counterfactual Edits for Generative Evaluation}

\author[1]{Maria Lymperaiou}[%
orcid=0000-0001-9442-4186,
email=marialymp@islab.ntua.gr]\cormark[1]
\author[1]{Giorgos Filandrianos}[%
email=geofila@islab.ntua.gr]\cormark[1]
\author[1]{Konstantinos Thomas}[%
orcid=0000-0002-7489-7776,
email=kthomas@islab.ntua.gr]\cormark[1]
\author[1]{Giorgos Stamou}[orcid= 0000-0003-1210-9874, email=gstam@cs.ntua.gr]
\address[1]{AILS Laboratory,
School of Electrical and Computer Engineering,
National Technical University of Athens}

\cortext[1]{Corresponding author.}

\begin{abstract}
Evaluation of generative models has been an underrepresented field despite the surge of generative architectures. Most recent models are evaluated upon rather obsolete metrics which suffer from robustness issues, while being unable to assess more aspects of visual quality, such as compositionality and logic of synthesis. At the same time, the explainability of generative models remains a limited, though important, research direction with several current attempts requiring access to the inner functionalities of generative models. Contrary to prior literature, we view generative models as a black box, and we propose a framework for the evaluation and explanation of synthesized results based on concepts instead of pixels. Our framework exploits knowledge-based counterfactual edits that underline which objects or attributes should be inserted, removed, or replaced from generated images to approach their ground truth conditioning. Moreover, global explanations produced by accumulating local edits can also reveal what concepts a model cannot generate in total. The application of our framework on various models designed for the challenging tasks of Story Visualization and Scene Synthesis verifies the power of our approach in the model-agnostic setting.
\end{abstract}


\maketitle

\section{Introduction}
Image generation has been one of the most popular deep learning tasks, inspiring many impressive state-of-the-art applications \cite{diff, diffusion, dalle, dalle2, imagen, semi, cc}. Even since the introduction of Generative Adversarial Networks (GANs) \cite{goodfellow2014gans}, which marked one of the first significant breakthroughs in the field, most applications focused on enhancing image quality according to human perception. 
At the same time, the automatic evaluation of the generated samples remains a long-standing problem as there are no ground truth data to measure against.
The valuation of such generative tasks, so far, relies on pixel-level metrics such as Inception Score (IS) \cite{is}, Frechet Inception Distance (FID) \cite{fid}, Learned Perceptual Image Patch Similarity (LPIPS) \cite{lpips}, to provide a quality measure for the generated samples. Consequently, the list of literature evaluated upon those benchmark metrics is long; yet concerns have been raised that their brittleness \cite{cleanfid} is leading to inaccurate results. Although recent metrics, such as Clean-FID \cite{cleanfid} can resolve some issues regarding visual artifacts, they still cannot address major issues such as the evaluation of complex images, compositionality, logic, and fairness of generation \cite{eval}. Moreover, when it comes to conditional generation, we further require a measure of whether objects and attributes mentioned in the conditioning are successfully depicted on the generated samples. Current attempts in conditional synthesis evaluation remain limited \cite{c-fid, c-eval} while still facing the shortcomings of their unconditional counterparts, on which they are built.

Explainability of generative models is another emerging field, which has currently received way less attention compared to discriminative models \cite{xaivision, xaivision2}. The incorporation of explainable feedback in Generative Adversarial Networks (GANs) has demonstrated a promising research direction \cite{xaigan}, while other works focus on interpreting GAN inner structure \cite{xaigan2}. Overfitting in GANs can be tackled by determining the image areas that contributed to classifying a sample as fake/real, thus explaining the discriminator's decision \cite{xaigan3}. This limited literature impedes the development of \textit{explainable evaluation} for generative models, even though related attempts have gained ground in other AI domains, such as Natural Language Processing \cite{nlpxai, nlpxai2, nlpxai3}.

We argue that resolving generative evaluation challenges calls for a conceptual approach to the evaluation process, diverging from the pixel-level route. Relying on concepts instead of pixels offers the advantage of enhanced interpretability regarding the evaluation process and paves the way for \textit{explainable evaluation} of generative models. Identifying concepts (objects or attributes) that can or cannot be generated reveals the capabilities and biases of the model at hand, thus driving potential architectural modifications.  In this paper, we present the first \textit{explainable evaluation} technique targeting generative models. Specifically, we utilize \textit{counterfactual explanations} to frame conditional generative evaluation as the answer to the following question: \textit{What concepts need to change in a generated sample X, for it to reach its conditioning c?} Conceptual edits guided from external knowledge sources \cite{cece} have shown to efficiently indicate the shortest possible path to reach the conditioning concepts. Furthermore, existing works that combine explainability with image generation operate on specific models \cite{xaigan, xaigan2, xaigan3} and demand access to their inner structure (white-box techniques), while our approach only requires generated outputs along with their ground truth conditioning, yet still regarding the generative model as a black-box.
We, therefore, contribute to the following:

\begin{enumerate}
    \item We propose the first \textit{conceptual} rather than pixel-based generative evaluation framework\footnote{https://github.com/geofila/Counterfactual-Edits-for-Generative-Evaluation}, suitable for various tasks such as Scene Generation (SG) and Story Visualization (SV).
    \item Our metrics are \textit{explainable} by design, illustrating which concepts need to be inserted, deleted, or replaced in the generated images, for them to approach the ground truth conditioning. Those edit operations are applied in a model-agnostic setting, totally trespassing any access to the generative model inner workings.
    \item Global explanations automatically reveal possible \textit{blind spots} of generative models, i.e. concepts that a model is intrinsically incapable of generating. 
\end{enumerate}

\section{Related work}
\paragraph{Generative Adversarial Networks (GANs)} \cite{goodfellow2014gans} consist of two neural networks, a generator $G(z; \theta_g)$ and a discriminator $D(x; \theta_d)$. $G$ maps random noise $z$, generated from a prior distribution $z \sim p_z$, to the data space. $D$, on the other hand, maps a sample $x_i$ from the same data space to a scalar value $p_i=D(x_i)$, which represents the probability that the sample was drawn from the real data distribution.
In the case of conditional GANs (cGANs) \cite{mirza2014condgan}, $G$ is fed not only with random noise $z$, but also with an additional conditioning vector $y$, which helps guide the generation of samples from specific sub-regions of the target distribution.

Several image generation cGANs \cite{odena2017conditional, miyato2018cgans} perform well when it comes to generating images with distinct textures and colors. However, they tend to struggle with generating coherent overall object structures and other long-range dependencies, due to the limited nature of convolutional filters. The Self-Attention GAN (SAGAN) \cite{zhang2019sagan} was proposed as a solution to this problem; it utilizes a self-attention module in both $G$ and $D$, as well as modern stabilization techniques such as Spectral Normalization of weights \cite{miyato2018spectral}, while it leverages the two-timescale update rule \cite{heusel2018ttur} to impose different learning rates for $G$ and $D$.

\paragraph{Diffusion models} are breaking new ground in the field of conditional image generation and are becoming the state-of-the-art in that area \cite{diffusion}. These models work by adding noise to an image and then learning to reconstruct it. In the past year, there have been several exciting developments in the field of diffusion-based image synthesis.
Stable Diffusion \cite{diff0} allows for high-quality image synthesis even under resource constraints by applying the diffusion process in the latent space of autoencoders instead of at the pixel level in the image space.
DALL-E2 \cite{dalle2} builds upon the success of its predecessor \cite{dalle} by incorporating text-conditioned image embeddings learned from CLIP \cite{clip} as input to a diffusion model that acts as the decoder. The resulting images are photorealistic and accurate representations of the input text, and it also allows for language-guided manipulation of a source image.
The work of Imagen \cite{diff1} leverages large pre-trained language models, such as T5 \cite{t5}, for language encoding and conducts image synthesis using the diffusion process.
DreamBooth \cite{diff2} takes Imagen a step further by allowing for context-aware image synthesis, given a text description of the desired context. This allows for the generation of various visual subjects while maintaining high image synthesis quality.

\paragraph{Conditional image synthesis} has come a long way since the early days of text-conditioned image generation \cite{reed2016t2i,reed2016t2i2}. First attempts produced images lacking in detail and quality. StackGAN \cite{stackgan} was the first model to significantly improve the quality of generated images using a multi-stage adversarial training process, followed by StackGAN++ \cite{zhang2018stackgan} which further enhanced generation results. AttnGAN \cite{xu2018attngan} employed attention mechanisms to generate fine-grained details in images based on individual words in the input text. SEGAN \cite{segan} took this a step further by only focusing attention on relevant keywords in the input text. DM-GAN \cite{zhu2019dmgan} improved the quality of generated images by addressing fuzzy areas.

StoryGAN \cite{storygan} is a generative model that synthesizes images based on sequential input (Story Visualization), using an RNN structure to encode the input text and provide context information to the generator. The generator is trained adversarially against two discriminators: the image discriminator, which evaluates image quality and text-image relevance, and the story discriminator, which ensures consistency across images given the entire story context. Recent work has focused on improving the baseline StoryGAN model \cite{li2020storygan} and exploring alternative story encoding methods, such as using Transformer architectures \cite{Maharana2021ImprovingGA,Maharana2021IntegratingVL, impartial}.

\paragraph{Generative evaluation}
Despite the rapid advancements in image synthesis, generative evaluation is falling behind due to outdated evaluation practices \cite{is, fid, lpips, psnr}, mainly followed for benchmarking reasons, ignoring the problems recognized in recent works \cite{cleanfid, eval}. Explainability in generative modeling can deliver interesting insights, though current efforts either remain model-specific \cite{gan-cannot, xaigan, xaigan2, xaigan3} or require discovering interpretable latent directions \cite{face, interface, compositionality, diamond}, which is a non-trivial task. 
Our method serves both the evaluation and explainability of generative models under a single framework and is capable of being adapted to any generative model - even the ones serving sequential image generation \cite{storygan, Maharana2021ImprovingGA, Maharana2021IntegratingVL, impartial} - as it focuses solely on input and output concept sets.
\paragraph{Counterfactual explanations}
Contemporary AI research moves towards explaining a neural network's train of 'thought', thus eXplainable AI (XAI) becomes a field of increasing interest \cite{xai}. Counterfactual explanations provide alternative realities based on minimal input modifications, hence revealing reasoning paths. Generative models are a straightforward approach when visual alternatives are explored \cite{ijcai2020}. Any alteration should be feasible with respect to original data distribution, an observation that adds constraints in alternative inputs \cite{Poyiadzi2020FACEFA}. Minimum alterations can be decided either by interfering with the black-box nature of neural networks \cite{cve}, or not \cite{cece}. We chose to follow the black-box route, using counterfactual explanations to uncover clues on the reasoning processes of generative models.

\section{Conceptual edits as counterfactual explanations}
\label{sec:edits}

Our overall approach is heavily inspired by \citet{cece}, which explores the fundamental question of counterfactual reasoning: “What is the \textit{minimal} change that has to occur in order for an image $I$ to be classified as X instead of Y?”, where X and Y are predicted categories of a pre-defined image classifier $F$. In our case, $F$ is not necessary, since we by default place all generated concepts $s$ in a set $S$, and all ground truth concepts $t$ in a set $T$. Counterfactual explanations are capable of addressing the aforementioned question, providing the minimum number of conceptual edits to achieve the $S \rightarrow T$ transition for all $s\in S$, $t \in T$. 

\textbf{Concept distances} instruct the shortest path that connects two specific concepts. Concept hierarchies are employed, deterministically defining the transition cost between concepts. We explore both the option to use external hierarchical knowledge such as WordNet \cite{wordnet}, mapping extracted concepts to synsets, or alternatively handcraft specific hierarchies to allow highly controlled semantic distance definition. In both cases, we denote as $d(s, t)$ the distance between concepts $s$ and $t$.
There are three available concept \textbf{edit operations} to realize transitions:
\begin{itemize}
    \item \textbf{Replacement (R)} $e_{s \rightarrow t}(S)$: A concept $s \in S$ is replaced with a concept $t  \notin S$.
    \item \textbf{Deletion (D)} $e_{s-}(S)$: A concept $s \in S$ is deleted from $S$.
    \item \textbf{Insertion (I)} $e_{s+}(S)$: A concept $s \in S$ is inserted in $S$.
\end{itemize}

Each edit operation inherits the concept distances imposed by the selected hierarchy. Therefore, \textbf{R} operation considers the path between $s$ and $t$ so that $\textit{min}(d(s, t))$ is ensured. As in \cite{cece}, $d$ also ensures \textit{actionability} of edits, allowing semantically meaningful transitions (e.g. 'food'$\rightarrow$'pasta'), while prohibiting meaningless ones (e.g. 'food'$\rightarrow$'sky').
\textbf{D} and \textbf{I} operations regard the \textit{root node} of the hierarchy as $t$ and $s$ respectively; 
in the case of WordNet, \textsf{entity.n.01} serves as the root. 
\textbf{Concept Set Edit Distance (CSED)} $D(S \rightarrow T)$ is obtained by aggregating all possible minimum cost edit operations so that $S \rightarrow T$ is finally achieved:  
\begin{equation}
  CSED = D(S \rightarrow T) = \textit{min}\sum_{s \neq t}^{S, T}\sum^{R, D, I} d(s, t)
\end{equation}


\section{Method}
\begin{figure}[t!]
    \centering
    \vspace{-18px}
    \includegraphics[width=0.95\textwidth]{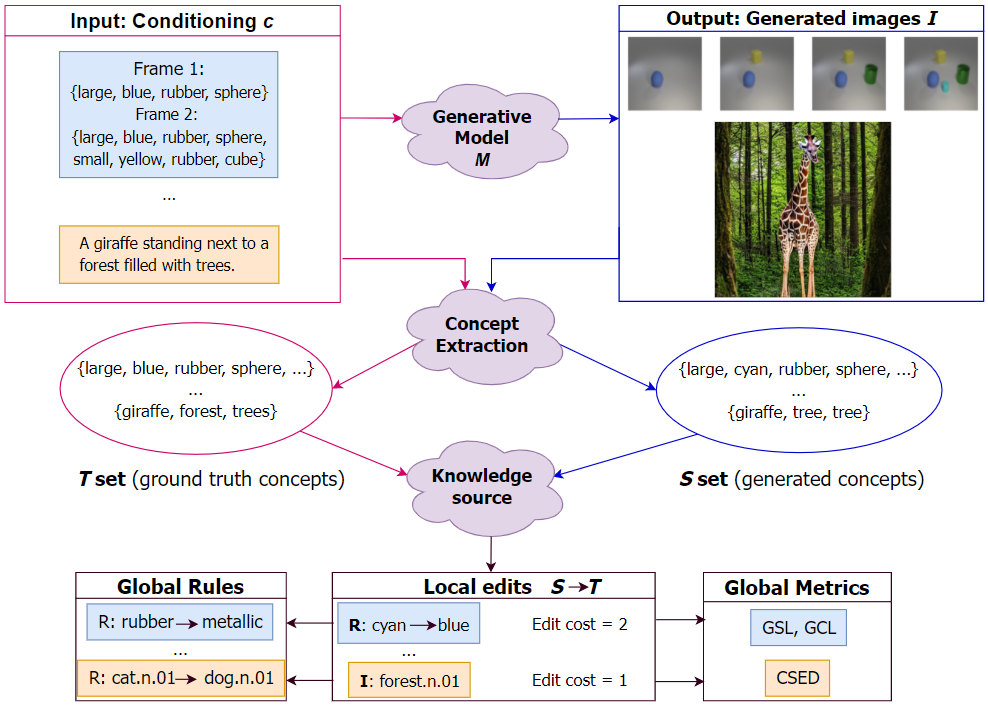}
    \caption{Outline of the proposed framework.}
    \label{fig:outline}
\end{figure}

The heart of our method consists of a pre-trained black-box generative model $M$ which receives a semantic description $c$ (in natural language or in symbolic format) as conditioning and produces an image $I$ corresponding to $c$. We then use off-the-self automatic methods such as object detection, semantic segmentation, and others, in order to extract all the concepts depicted in the generated images $I$ and append them in the \textit{generated} (or \textit{source}) concept set $S$. Similarly, concepts extracted from $c$ contribute to the \textit{real} (or \textit{target}) concept set $T$. The format of $c$ defines the concept extraction technique that is followed, ranging from linguistic concept extraction, if $c$ is a textual sentence, to simple preprocessing, if $c$ is already in a set format. Ultimately, we aspire to answer the following: "What are the \textit{minimal} required changes in order to traverse from $S$ to $T$?" The outline of our method is presented in Figure \ref{fig:outline}.

\subsection{Generative evaluation}
The counterfactual backbone described in Section \ref{sec:edits} highlights our process for generative evaluation, which we employ on two difficult tasks of the generative literature: \textbf{Story Visualization (SV)} and \textbf{Scene Generation (SG)}.

\paragraph{Story Visualization (SV)} targets the sequential creation of images $I_1, I_2, ..., I_L$ that correspond, one-to-one, to a given sequential conditioning $c_1, c_2, ..., c_L$ of a total length $L$. The generated images need not only to remain \textit{faithful} to their conditioning, but to also maintain serial \textit{consistency}. We therefore define the two desiderata applicable to SV:

\begin{itemize}
    \item \textbf{Faithfulness}: objects and attributes mentioned in $c_k$ should also appear in frame $I_k$, for any story frame $k$, where $1\leq k \leq L$.
    \item \textbf{Consistency}: objects or attributes appearing in frame $I_k$ cannot disappear or change in later frames $I_{k+1}, ..., I_L$, for any story frame $k$, where $1\leq k < L$.
\end{itemize}

Since well-defined semantics are tied to counterfactual explanations \cite{counterfactual1}, we regard CLEVR-SV \cite{clevr} as the ideal dataset to demonstrate our approach, as it provides a set of concepts $\mathcal{C}$: shape \textit{(cube, sphere, cylinder)}, size \textit{(small, large)}, material \textit{(rubber, metal)} and one of 8 colors \textit{(blue, cyan, brown, yellow, red, green, purple, gray)}. Each CLEVR-SV object contains $|\mathcal{C}|$=4 concepts that describe its shape, size, material and color. We handcraft a simple hierarchy to group object semantics to generic concept classes, demonstrating the following inclusion relationships:
\begin{equation}
\label{eq:hierarchy}
\begin{split}
\textit{(large, small)} \subset \textit{Size} \\
\textit{(blue, yellow, brown, grey, green, purple, cyan, red)}\subset \textit{Color} \\
\textit{(metallic, rubber)} \subset \textit{Material} \\
\textit{(sphere, cube, cylinder)} \subset \textit{Shape}   
\end{split} 
\end{equation}
CLEVR-SV contains stories of length $L$=4, with the $k$-th frame strictly containing $k$ objects.
Any of the three available edit operations can be relevant per frame: \textbf{D} of a concept, when a generated frame contains more objects than its ground truth match; \textbf{I} of a concept in the opposite case; \textbf{R} equals to a \textbf{D} followed by an \textbf{I}, and can be applied on frames with proper number of objects when semantics differ. In the default case, we assign equal costs of 1 for all semantics, as well as for \textbf{D} and \textbf{I} operations (\textbf{R} cost is the sum of \textbf{D} and \textbf{I} costs).

To measure story \textbf{faithfulness} we propose the \textbf{\textit{Story Loss (SL)}} metric, which sums up the per-frame Concept Set Edit Distance ($CSED_k$) for $k=1, 2, ..., L$ frames of the story. Generated CLEVR-SV semantics for shape, size, material and color for the \textit{k}-th frame form the concepts set $S_k$, while the semantics of the conditioning form $T_k$, with their $CSED_k$ denoted as $D(S_k, T_k)$. Thus, the cost for the transition $\{S_1, S_2, S_3, S_4\}\rightarrow \{T_1, T_2, T_3, T_4\}$ corresponding to the minimum cost \textbf{R} \textbf{D} \textbf{I} edits needed to transform the semantics of the generated sequence $\{I_1, I_2, I_3, I_4\}$ to the semantics of its conditioning $\{c_1, c_2, c_3, c_4\}$ can be expressed as:
\begin{equation}
  SL =  \sum_{k=0}^{k=L} CSED_k = \sum_{k=0}^{k=L} D(S_k, T_k), \qquad L=4
\end{equation}
By scaling up the calculation of SL for a dataset containing $N$ stories, we obtain the \textbf{\textit{Global Story Loss (GSL)}} metric:
\begin{equation}
  GSL =  \sum_{i=0}^{N} SL_i = \sum_{i=0}^{N}\sum_{k=1}^{k=L} D(S_k, T_k)_i
\end{equation}
As for story \textbf{consistency}, we propose the metric of \textbf{\textit{Consistency Loss (CL)}}: the frame
$I_k$ is compared with $I_{k-1}, k=2, .., L$ frames of the generated sequences to capture changes of semantics. A challenging aspect of CL is that there does not exist a ground truth concept set. However, since it is known by task definition that the \textit{k}-th frame contains \textit{k} objects, and the cardinality $|\mathcal{C}|$ of dataset concepts is predefined ($|\mathcal{C}|$=4 in the case of CLEVR-SV), we can assume that every previous frame constitutes the 'ground truth' corresponding to the concept set $T$. Commencing from the \textit{k}=1 frame, we expect the cardinality of $T$ to be equal with $|\mathcal{C}|\cdot k = |\mathcal{C}|$. Any discrepancy results in a penalty $p_k = |T|- |\mathcal{C}|\cdot k = CL_k$ for \textit{k}=1. For later frames, we define as $S$ the concept set corresponding to the \textit{k}-th frame, and as $T$ the 'ground truth' set comprised of the \textit{k}-1 frame concepts. Mathematically, CL can be written as:
\begin{equation}
\label{eq:cl}
  CL = p_{k=1} + \sum_{k=2}^{k=L} D(S_k, T_k), \qquad T_k = S_{k-1}, L=4
\end{equation}
In the ideal case, when the \textit{k}-th frame contains \textit{k} objects with $\mathcal{C}$ semantics, we expect that $p_{k=1}=0$ and $CL_{k>1}=|\mathcal{C}| \cdot (k-1)$.
By extending CL to $N$ stories, \textbf{\textit{Global Consistency Loss (GSL)}} evaluates the consistency capabilities of a generative model $M$ in total:
\begin{equation}
  GCL =  \sum_{i=0}^{N} CL_i = \sum_{i=0}^{N} \{p_{k=1, i} + \sum_{k=2}^{k=L} D(S_k, T_k)_i, \;\;  T_{k, i} = S_{k-1, i}\}
\end{equation}
Average values can be obtained for both local (\textbf{SL}/\textbf{CL}) and global (\textbf{GSL}/\textbf{GCL}) metrics:
\begin{equation} \label{eq:avgsl}
    \textit{Avg}\ SL = \frac{1}{k}SL, \qquad \textit{Avg}\ GSL = \frac{1}{N}[\textit{Avg}\ SL] =  \frac{1}{N}GSL
\end{equation}
For consistency, instead of exporting an average value over $\displaystyle \sum CL_{k}$, it is more meaningful to count how
many times the $p_{k=1}=0, CL_{k>1}=|\mathcal{C}| \cdot (k-1)$ requirement was not respected, averaged for $L=k$ frames:
\begin{equation} \label{eq:avgcl}
\textit{Avg}\ CL = \frac{p_{k=1}}{k} + \frac{1}{k} \sum_{k=1}^{k=L}[CL_{k>1}\neq|\mathcal{C}| \cdot (k-1)], \quad \textit{Avg}\ GCL = \frac{1}{N}[\textit{Avg}\ CL]
\end{equation}
\textbf{SL} and \textbf{CL} are by nature \textit{explanaible}, as they do not only provide a measure of quality but also reveal the $S_k \rightarrow T_k$ edit paths. Those paths serve as \textit{local counterfactual explanations}, highlighting the erroneously generated semantics for this particular story, either in terms of \textbf{faithfulness} or \textbf{consistency}. 
Overall, higher \textbf{SL}/\textbf{GSL} and \textbf{CL}/\textbf{GCL} values denote lower conceptual generation quality.
\textbf{GSL}/\textbf{GCL} edit paths correspond to \textit{global counterfactual explanations}: rule extraction techniques provide frequent patterns, summarizing the behavior of $M$ under investigation. 
Frequent \textbf{GSL} edit paths in fact contain \textit{common misconceptions}, i.e. conditioning concepts that $M$ cannot easily generate. Similarly, \textbf{GCL} edit paths reveal frequent \textit{inconsistency patterns}, showcasing concepts that arbitrarily change within the story frames.
Hence, by researching the question "What has to \textit{minimally} change in order to transit from $S$ to $T$?", we eventually answer a more generic one: "Which concepts cannot be generated or preserved by $M$?" 

\paragraph{Scene Generation (SG)} \label{sg} aims to synthesize a visual scene $I$ based on a conditioning $c$. The synthesized image comprises multiple objects which interact with each other. Scene objects are also accompanied by attributes. The given conditioning $c$ is more complex compared to conditionings provided for SV, since the concepts to be generated are numerous and not predefined; this yields a concept set $\mathcal{C}$ of unknown but comparatively large cardinality. 

COCO dataset \cite{coco} provides the ideal setting for evaluating generative \textbf{faithfulness} for SG, providing textual captions $c$ that can serve as conditioning. We focus our endeavors on state-of-the-art open source diffusion models \cite{diffusion} from Huggingface\footnote{\href{https://huggingface.co/models?pipeline_tag=text-to-image&sort=downloads}{https://huggingface.co/models?pipeline\_tag=text-to-image\&sort=downloads}}, and specifically on Stable Diffusion  v1.4 \& v2 \cite{stable1, stable2} and Protogen x3.4 \& 5.8 \cite{protogen, protogen58} (details in Appendix). These models produce \textit{realistic} images - an important aspect of the concept extraction (object detection) stage. We omit older SG architectures \cite[][inter alia]{spade, g2i19, g2i17, g2i16, g2i15} due to their inferior visual quality and their reliance on scene graphs and layouts for ensuring proper composition.

In the concept extraction stage, YOLO-v8 \cite{yolov8} and YOLOS \cite{yolos} object detectors are leveraged to construct the generated concept set $S$. Since $c$ is in textual format, spaCy \cite{spacy} is used to extract \textit{ground truth concepts} from captions that form the target concept set $T$. 
The semantically complex nature of concept distances related to COCO concepts requires a rich knowledge scheme, such as WordNet. For example, if $c$ refers to concepts such as 'food' or 'animal', a diffusion model may generate more refined 'food' or 'animal' instances, for example, 'pasta' and 'dog' respectively. The object detectors will then return these refined classes, inducing some noise in the transformation process. Hierarchical knowledge can eliminate such issues: even though $T=\{\textit{food}, \textit{animal}\}\neq S=\{\textit{pasta}, \textit{dog}\}$, the two sets are \textbf{semantically equivalent} if we consider the hierarchical relationships $\textit{pasta}-\textit{isA}-\textit{food}$ and $\textit{dog}-\textit{isA}-\textit{animal}$ provided by  mapping $S$ and $T$ concepts on WordNet synsets. In this case, no $S\rightarrow T$ transformation needs to be performed. Therefore, the usage of external knowledge allows more conceptually accurate transitions.
Moreover, WordNet provides concept distances necessary for edit operations, precisely reflecting semantic relationships between concepts. Then, CSED can be applied to provide the total cost of the $S\rightarrow T$ transformations.

\section{Experiments}
\subsection{Story Visualization}
Since all semantics and \textbf{D, I} edit operations have an equal cost, we assign $d$=1 for all semantics, as well as for \textbf{D, I}. For example, deleting a color yields an edit cost of 1. Alternatively, by substituting a color with another one induces an edit cost of 2, equal to deleting the source color and then inserting the target color. The same logic applies to shape, size and material of objects.
\paragraph{Metric results} over the best variants of selected SV models \cite{storygan, Maharana2021ImprovingGA, Maharana2021IntegratingVL, impartial} are presented in Table \ref{tab:all-results}. Existing metrics (FID, Clean-FID, LPIPS, SSIM) are provided for comparison. 

In general, we observe an agreement between pixel-level and conceptual metrics. This is somehow expected, since the concept extraction stage depends on pixel-level image quality, with better generated objects or semantics being more easily identifiable. Nevertheless, conceptual evaluation offers more \textit{explainable insights}: percentages of losses per concept (Material, Size, Shape, Color) are provided, highlighting strengths and shortcomings of investigated models over different semantics. For example, higher Shape loss for all models (> 50\%), indicates that they  synthesize objects of ambiguous shapes in most cases. On the other hand, relatively lower Size losses reveal the models' capability to generate objects having the right size.

\begin{table*}[h!]
\hspace{-20px}
\caption{Average evaluation metrics (existing and proposed, separated by vertical line) for all \textit{L}=4 stories per $M$.}
\label{tab:all-results}
\begin{tabular}{p{15pt}|p{50pt}p{24pt}p{45pt}p{20pt}|p{18pt}p{19pt}p{26pt}p{24pt}p{24pt}p{22pt}}
\toprule
$M$ & \centering FID& \centering Clean & \centering LPIPS & \centering SSIM & \centering GCL & \centering GSL & \centering Material & \centering Size  & \centering Shape & Color\\
&\centering $\downarrow$&-FID$\downarrow$&\centering $\downarrow$&\centering $\uparrow$&\centering $\downarrow$&\centering $\downarrow$&\centering $\downarrow$&\centering $\downarrow$&\centering $\downarrow$&$\;\; \;\downarrow$ \\
\midrule
\cite{impartial} & 41.54 $\pm$ 8.55 & \textbf{115.46} & \textbf{0.21} $\pm$ 0.05 & \textbf{0.71} & \textbf{4.97}&\textbf{7.01} & \textbf{20.83}\% & \textbf{14.55}\% & \textbf{56.62}\% & \textbf{33.10}\% \\
\cite{storygan} & \textbf{41.45} $\pm$ 6.25 & 123.40 & 0.25 $\pm$ 0.03 & 0.65 & 11.44 & 15.33  & 30.89\% & 21.12\% & 62.34\% & 37.44\%\\
\cite{Maharana2021IntegratingVL} & 41.96 $\pm$ 9.66 &  124.97 &  0.25 $\pm$ 0.08 &  0.67 & 10.95 & 8.06 & 21.45\% & 16.02\% & 56.78\% & 35.10\% \\
\cite{Maharana2021ImprovingGA}  & 41.80 $\pm$ 8.81 &  122.62 &  0.25 $\pm$ 0.05 &  0.68 & 8.32 & 11.51 & 25.34\%&  16.71\%& 56.83\%& 35.14\%\\
\bottomrule
\end{tabular}
\end{table*}

We further investigate our findings by focusing on the best performing SV model of \citet{impartial} according to the conceptual metrics reported in Table \ref{tab:all-results}. Specifically, in Table \ref{tab:frame-results} we present results of per frame \textbf{GSL}, \textbf{GCL} and losses per concept (Material, Size, Shape, Color).

\begin{figure*}[h!]
    \centering
    \begin{subfigure}{0.49\textwidth}
        \includegraphics[width=0.96\columnwidth]{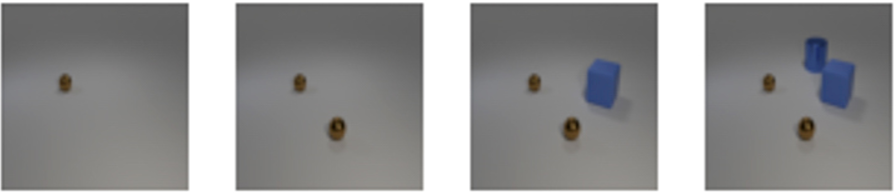}
        \caption{Ground truth story frames}
        \label{fig:a}
    \end{subfigure}
    \begin{subfigure}{0.49\textwidth}
        \includegraphics[width=0.96\columnwidth]{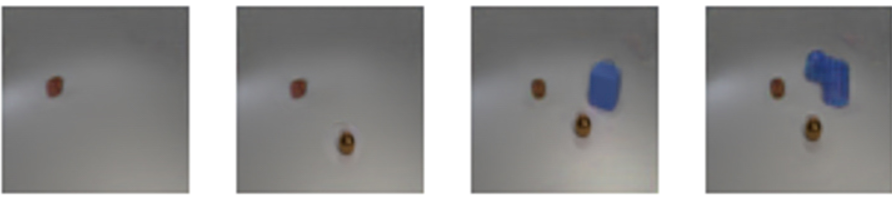}
        \caption{Generated story frames of \cite{impartial}}
        \label{fig:b}
    \end{subfigure}
    \caption{Ground truth vs generated CLEVR-SV story frames using \cite{impartial} for \textit{L}=4.}
    \label{fig:examples}
\end{figure*}

\begin{table*}[h!]
\caption{Average conceptual evaluation metrics per frame for \cite{impartial}.}
\label{tab:frame-results}
\begin{tabular}{c|p{18pt}p{19pt}cccc}
\toprule
Frame & GSC$\downarrow$& GSL$\downarrow$ & Material$\downarrow$  & Size$\downarrow$  & Shape$\downarrow$  & Color$\downarrow$ \\
\midrule
1st & 0.00 & 2.25 & 40.00\% & 6.20\% & 58.75\% & 7.50\% \\
2nd & 4.35 & 5.66 & 20.00\% & 11.88\% & 57.5\% & 32.50\% \\
3rd & 7.12 & 8.25 & 13.33\% & 16.67\% & 57.08\% & 43.33\%\\
4th & 8.42 & 11.49 & 10.00\% & 23.44\% & 53.13\% & 49.06\% \\
\bottomrule
\end{tabular}
\end{table*}
\paragraph{Local explanations}
The transparency of the proposed \textbf{SL}/\textbf{CL} metrics is verified by obtaining \textit{local explanations} for \cite{impartial}. Specifically, we examine edit paths for the sequences of Figure \ref{fig:examples}: the 4 leftmost images (Figure \ref{fig:a}) correspond to the ground truth sequence, while the 4 rightmost images (Figure \ref{fig:b}) denote the generated frames. Consequently, $S$ contains concepts of \ref{fig:b} and $T$ contains concepts of \ref{fig:a}.
As presented in Table \ref{tab:explain} (details in Appendix), a standard \textbf{R} operation for all frames is observed, suggesting transforming the material of the small brown sphere from \textit{'rubber'} to \textit{'metallic'} in order to match the ground truth. In the last frame, one more \textbf{R} operation is added, suggesting also transforming the shape of the new object from \textit{'sphere}' to \textit{'cylinder'}. The cost for each \textbf{R} operation equals to 2, equivalent for one step to remove the wrong semantic and one more step to add the right semantic. However, this cost weight can be tuned appropriately, if needed. SL for this story equals to 10, as a summary of all operation costs per frame. 
By observing for CL, we realize that the correct number of objects is added in every consequent frame, so that CL$_{k>1}=|\mathcal{C}| \cdot (k-1), |\mathcal{C}|=4$ is maintained: starting from CL$_1=p_{k=1}=0$ for the \textit{k}=1 frame, we verify that only one object is added, respecting that frame number should be equal to the number of objects present in it. CL$_2$=4 is expected since the object added in the \textit{k}=2 frame contains 4 semantics. Any number lower or greater than that would indicate an abnormal behavior: CL$_{k>1}<|\mathcal{C}| \cdot (k-1)$ marks one (or more) missing objects, while CL$_{k>1}>|\mathcal{C}| \cdot (k-1)$ indicates one (or more) extra object generated. The desired pattern repeats for the 3rd and 4th frames.
Through this analysis, the shortcomings of \cite{impartial} concerning this specific image are revealed, producing a \textit{local} explanation: The semantic \textit{Material} needs to be examined more, as in all story frames of this example the small brown sphere is generated with the attribute \textit{'rubber'} instead of \textit{'metallic'}. In order to obtain insights regarding the model's synthesis capabilities of discrete semantics, global metrics and explanations need to be derived.

\begin{table}[h!]
\caption{Interpretable local edits for Figure \ref{fig:examples}.}
\label{tab:explain}
\begin{tabular}{p{13pt}ccp{35pt}cp{14pt}}
\toprule
Frame  & Min edit path  & Operation & Edit cost &  Semantic & CL \\
\midrule
1st & 'rubber' $\rightarrow$'metallic' & \textbf{R}  & \;\;\; 2 & Material & \;\;0 \\
2nd & 'rubber' $\rightarrow$'metallic' & \textbf{R}  & \;\;\; 2 & Material & \;\;4 \\
3rd & 'rubber' $\rightarrow$'metallic' & \textbf{R}  & \;\;\; 2 & Material & \;\;8 \\
4th &  \{'rubber', 'sphere'\} $\rightarrow$ \{'metallic', 'cylinder'\} & \textbf{R}, \textbf{R}   & \;\;\; 4 & Material, Shape & \;12 \\
\bottomrule
\end{tabular}
\end{table}

\paragraph{Global explanations}
In order to assess our model's shortcomings in total, we measure \textbf{GSL} for all test images of CLEVR-SV. Therefore, we can obtain a measure of the model's inability to capture certain -discrete- semantics, either per frame or in total (Table \ref{tab:all-results}). We observe that in later frames, \textit{Material loss} decreases, even though we would expect that the problem gets harder and harder as more objects are added, resulting in higher losses. This expected pattern is followed by \textit{Size loss} and \textit{Color loss}, while no certain pattern can be extracted from \textit{Shape loss}. 
The high \textit{Shape loss} imposes the need for attention mechanisms within the used GANs \cite{zhang2019sagan}, so that long-range relationships can be captured. We can also attribute the rapid rise of \textit{Size} and \textit{Color} losses to consistency deficiencies within the story sequence. 

\textbf{GSL} can also reveal patterns in the form of rules for the whole test set. We leverage the apriori algorithm \cite{apriori} to extract frequent semantic combinations and rules. The 4 most common semantic edits are provided in Table \ref{tab:global_rules}, together with each rule's frequency (support). The concept category (as occurring from equation \ref{eq:hierarchy}), antecedent support (source semantic frequency), and consequent support (target semantic frequency) are also provided.

We observe that \textit{Material} is the most common concept misconception, with both 'rubber' and 'metallic' semantics being frequently confused. \textit{Shape} is the second most prominent misconception, with 'cylinder' appearing in the generated frames more often compared to the 'cylinder' occurrence in the conditioning; 'cube' and 'sphere' shapes are sacrificed for 'cylinder' to be generated. Since the rule support is not significantly high, with 26.77\% being the maximum value, we can safely assume that the SV model of \citet{impartial} is not heavily biased towards certain semantics. Nevertheless, we spot some tendency to generate the wrong material and shape, an observation that can be valuable for architectural improvements of the model. 

\begin{table}[t!]
\caption{Interpretable global edits on test set images of CLEVR-SV generated using \cite{impartial}.}
\label{tab:global_rules}
\begin{tabular}{ccccc}
\toprule
Rules (edits) & Semantic  & Support \%  &  Antec. support\%& Conseq. support\% \\
\midrule
'metallic' $\rightarrow$'rubber' &  Material & 26.77 & 26.77 & 26.77\\
'rubber' $\rightarrow$'metallic' & Material & 22.05 & 22.05 & 22.83\\
'cylinder' $\rightarrow$'cube' & Shape &  18.11 & 33.07 & 31.50\\
'cylinder' $\rightarrow$'sphere' & Shape & 14.96 & 33.07 & 18.90\\
\bottomrule
\end{tabular}
\end{table}

\subsection{Scene Generation}
We select the first 10K samples from COCO to reduce the inference time needed to extract visual concepts using YOLO-v8 and YOLOS object detectors. COCO provides 5 descriptive sentences per sample, which are paraphrases of each other. For this reason, we only regard the 1st out of the 5 sentences as the conditioning $c$. We follow two separate processes for SG: actual \textit{generation} conditioned on $c$ and \textit{retrieval} of caption-image pairs based on captions similar to $c$.

\paragraph{Conditional generation on COCO captions}
For the generation experiment, we employ pre-trained diffusion models without any further tuning, as mentioned in \ref{sg}, which are all tested on the same conditionings $c$. Each of the four diffusion models required about 15 hours to synthesize 10K images using 2 T4 GPUs, therefore around 60 hours in total. 

\paragraph{Retrieval of COCO-related captions}
In order to obtain considerably more images conditioned on COCO-related queries without having to spend the time and resources to run many more thousand iterations of the diffusion model, we utilized a Stable Diffusion search engine (Lexica.art)\footnote{\href{https://lexica.art/}{https://lexica.art/}}. 
The exact process we used was the following: we use $c$ of the first 10K COCO samples as the 'query' caption. The search engine returned, for each of the 10k captions, 10 images that have been already generated by online communities with the closest input queries to our captions. This technique supplied us with 100.000 more Stable Diffusion images, accompanied by their input queries. We then compare results between web-retrieved and generated images.

\paragraph{Object detection}
We select a default threshold of $T_d=$0.6 for detection; objects detected with confidence$\geq$0.6 are added in the generated concept set $S$. This threshold is experimentally defined to maintain a valid trade-off between false positive and false negative objects; in fact, since no ground truth exists, even defining false predictions is untractable without human inspection. However, our approach can provide relevant hints regarding the probability of false detection, as a higher number of \textbf{D} operations may infer higher false positive rates (irrelevant objects being detected, if $T_d$ is too low), while more \textbf{I} operations can be correlated with higher false negative rates (relevant objects not being detected, if $T_d$ is too strict). 

\paragraph{Metric results}
For comparative reasons we present results for $T_d=$0.5, 0.6, 0.7 in Tables \ref{tab:diffusion-yolov8} (YOLO-v8) \& \ref{tab:diffusion-yolovs} (YOLOS) for generated images, and in Table \ref{tab:scrapped} for web images, reporting object extraction from both object detectors. Instances colored in \textcolor{blue}{blue} denote the lowest scores, which are more desirable, while the highest scores are highlighted with \textcolor{red}{red}. We present number of edits (\# \textbf{I}, \# \textbf{D}, \# \textbf{R}), as well as the total cost for each \textbf{I}, \textbf{D}, \textbf{R} operation for all images. Mean CSED is reported as an overall metric regardless of which operation was performed more often.

\begin{table}[htp!]
\caption{Metric results using YOLO-v8 for object detection on generated images from COCO queries.}
\label{tab:diffusion-yolov8}
\begin{tabular}{c|>{\centering\arraybackslash}p{7em}|>{\centering\arraybackslash}p{3em}>{\centering\arraybackslash}p{3em}>{\centering\arraybackslash}p{3em}>{\centering\arraybackslash}p{3em}>{\centering\arraybackslash}p{3em}>{\centering\arraybackslash}p{3em}>{\centering\arraybackslash}p{5em}}
\toprule
$T_d$  & $M$ & \# \textbf{I} & Cost \textbf{I} & \# \textbf{D} & Cost \textbf{D} & \# \textbf{R} & Cost \textbf{R} & Mean CSED\\
\midrule
\multirow{3}{1em}{0.5} & stable diffusion & 37651 & 16762 & 1196 & 5655 & 126004 & 14323 & 35.75 \\
& stable diffusion 2 & \textcolor{blue}{36878} & \textcolor{blue}{16067} & \textcolor{red}{1243} & \textcolor{red}{6301} & \textcolor{red}{129315} & \textcolor{red}{14839} & \textcolor{red}{36.32} \\
& protogen base & 37072 & 16208 & 1233 & 5944 & 129290 & 14744 & 35.95 \\
& protogen 5.8 & \textcolor{red}{38581} & \textcolor{red}{17715} & \textcolor{blue}{1195} & \textcolor{blue}{4702} & \textcolor{blue}{117708} & \textcolor{blue}{13411} & \textcolor{blue}{34.66} \\
\midrule
\multirow{3}{1em}{0.6} & stable diffusion & 39070 & 18386 & 1157 & 4042 & 110260 & 12964 & 34.22 \\
& stable diffusion 2 & 38678 & \textcolor{blue}{17782} & \textcolor{red}{1200} & \textcolor{red}{4514} & 112499 & 13397 & \textcolor{red}{34.55} \\
& protogen base & \textcolor{blue}{38548} & 17794 & 1184 & 4270 & \textcolor{red}{114762} & \textcolor{red}{13427} & 34.35 \\
& protogen 5.8 & \textcolor{red}{39766} & \textcolor{red}{19210} & \textcolor{blue}{1134} & \textcolor{blue}{3419} & \textcolor{blue}{103579} & \textcolor{blue}{12135} & \textcolor{blue}{33.38} \\
\midrule
\multirow{3}{1em}{0.7} & stable diffusion & 40814 & 20391 & 1086 & 2681 & 93390 & 11337 & 32.96 \\
& stable diffusion 2 & 40677 & 19806 & \textcolor{red}{1107} & \textcolor{red}{2938} & 95477 & 11756 & \textcolor{red}{33.08} \\
& protogen base & \textcolor{blue}{40397} & \textcolor{blue}{19801} & 1101 & 2820 & \textcolor{red}{97314} & \textcolor{red}{11787} & 32.94 \\
& protogen 5.8 & \textcolor{red}{41295} & \textcolor{red}{20944} & \textcolor{blue}{1039} & \textcolor{blue}{2308} & \textcolor{blue}{89850} & \textcolor{blue}{10726} & \textcolor{blue}{32.39} \\
\bottomrule
\end{tabular}
\end{table}

\begin{table}[t!]
\caption{Metric results using YOLOS for object detection on generated images from COCO queries.}
\label{tab:diffusion-yolovs}
\begin{tabular}{c|>{\centering\arraybackslash}p{7em}|>{\centering\arraybackslash}p{3em}>{\centering\arraybackslash}p{3em}>{\centering\arraybackslash}p{3em}>{\centering\arraybackslash}p{3em}>{\centering\arraybackslash}p{3em}>{\centering\arraybackslash}p{3em}>{\centering\arraybackslash}p{5em}}
\toprule
$T_d$  & $M$ & \# \textbf{I} & Cost \textbf{I} & \# \textbf{D} & Cost \textbf{D} & \# \textbf{R} & Cost \textbf{R} & Mean CSED\\
\midrule
\multirow{3}{1em}{0.5} & stable diffusion & \textcolor{blue}{26302} & 9032 & 1382 & 44189 & \textcolor{red}{197623} & \textcolor{red}{21097} & 68.25 \\
& stable diffusion 2 & 26684 & \textcolor{blue}{8832} & 1403 & 43459 & 192198 & 21082 & 68.05 \\
& protogen base & 26887 & 8966 & \textcolor{red}{1404} & \textcolor{red}{44406} & 193327 & 21035 & \textcolor{red}{68.81} \\
& protogen 5.8 & \textcolor{red}{28880} & \textcolor{red}{10367} & \textcolor{blue}{1373} & \textcolor{blue}{34996} & \textcolor{blue}{189677} & \textcolor{blue}{19858} & \textcolor{blue}{60.45} \\
\midrule
\multirow{3}{1em}{0.6} & stable diffusion & \textcolor{blue}{27963} & 9920 & 1373 & 33891 & \textcolor{red}{188395} & 20286 & 60.10 \\
& stable diffusion 2 & 28145 & \textcolor{blue}{9662} & \textcolor{red}{1394} & 33933 & 182767 & \textcolor{red}{20322} & 60.36 \\
& protogen base & 28499 & 9845 & \textcolor{red}{1394} & \textcolor{red}{34167} & 185217 & 20224 & \textcolor{red}{60.63} \\
&  protogen 5.8 & \textcolor{red}{30545} & \textcolor{red}{11330} & \textcolor{blue}{1364} & \textcolor{blue}{27218} & \textcolor{blue}{179947} & \textcolor{blue}{18963} & \textcolor{blue}{54.13} \\
\midrule
\multirow{3}{1em}{0.7} & stable diffusion & \textcolor{red}{29998} & \textcolor{red}{10985} & 1357 & \textcolor{blue}{24956} & 177213 & \textcolor{blue}{19319} & \textcolor{blue}{52.51} \\
& stable diffusion 2 & \textcolor{blue}{29831} & 10657 & 1347 & 25492 & \textcolor{blue}{172860} & 19409 & 53.14 \\
& protogen base & 29866 & 10790 & \textcolor{blue}{1346} & 25255 & 175495 & 19350 & 52.98 \\
& protogen 5.8 & 28880 & \textcolor{blue}{10367} & \textcolor{red}{1373} & \textcolor{red}{34996} & \textcolor{red}{189677} & \textcolor{red}{19858} & \textcolor{red}{60.45} \\
\bottomrule
\end{tabular}
\end{table}

\begin{table}[htp!]
\caption{Metric results for web-retrieved Stable Diffusion images on similar queries to COCO.}
\label{tab:scrapped}
\begin{tabular}{c|>{\centering\arraybackslash}p{7em}|>{\centering\arraybackslash}p{3em}>{\centering\arraybackslash}p{3em}>{\centering\arraybackslash}p{3em}>{\centering\arraybackslash}p{3em}>{\centering\arraybackslash}p{3em}>{\centering\arraybackslash}p{3em}>{\centering\arraybackslash}p{5em}}
\toprule
$T_d$  & Obj. detector & \# \textbf{I} & Cost \textbf{I} & \# \textbf{D} & Cost \textbf{D} & \# \textbf{R} & Cost \textbf{R} & Mean CSED\\
\midrule
\multirow{2}{1em}{0.5} 
& YOLO-v8 & 186775 & 857448 & 1343 & 52247 & 1353479 & 224350 & 75.87 \\
 & YOLOS & 163628 & 605321 & 1487 & 421525 & 2469635 & 473331 & 106.41 \\
\midrule
\multirow{2}{1em}{0.6} 
& YOLO-v8 & 190047 & 891454 & 1317 & 37418 & 1174012 & 190928 & 73.74 \\
& YOLOS & 167576 & 646112 & 1467 & 308346 & 2303966 & 432851 & 98.06 \\
\midrule
\multirow{2}{1em}{0.7} & YOLO-v8  & 193663 & 929183 & 1236 & 25388 & 982259 & 154063 & 71.81 \\
& YOLOS & 171778 & 688942 & 1449 & 214928 & 2115779 & 390304 & 90.56 \\
\bottomrule
\end{tabular}
\end{table}

Regarding the selected threshold $T_d$, our initial hypothesis is proven to be correct: more \textbf{I} operations are realized for higher threshold $T_d$=0.7, suggesting that objects from the conditioning where not detected, while fewer \textbf{I} were performed for $T_d$=0.5. Similarly, there are more \textbf{D} operations for the lowest $T_d$=0.5, as spurious objects can be detected more easily. Additionally, more \textbf{R} operations are needed for lower thresholds, which is also expected, since more objects are extracted and added to the $S$ set. As for object detectors, results using YOLO-v8 are very homogeneous, indicating that the models under investigation follow a rather predictable behavior irrespectively of $T_d$.
Protogen 5.8 consistently yields the lowest mean CSED score, denoting cheaper transitions for all thresholds. This observation slightly changes for $T_d$=0.7 and YOLOS object detector (Table \ref{tab:diffusion-yolovs}), for which, surprisingly, protogen 5.8 produces the more expensive transitions. By comparing Tables \ref{tab:diffusion-yolov8} \& \ref{tab:diffusion-yolovs}, YOLOS results in higher mean CSED, less \textbf{I} operations, significantly more expensive \textbf{D} operations (even though the number of \textbf{D} operations is not substantially larger), as well as more and expensive \textbf{R} operations. Therefore, we can safely assume that YOLOS is comparatively more sensitive in detecting more objects, which may induce some noise in the detection process. 
All these results will become more interpretable should we delve into the explanations accompanying the evaluation.
The patterns arising from evaluating generated images are also supported in Table \ref{tab:scrapped} findings, verifying the threshold hypothesis, as well as the increased sensitivity of YOLOS. Nevertheless, web-retrieved images seem to miss objects mentioned in the query, as proven by the large number of \textbf{I} and \textbf{R} operations.

\begin{figure}[h!]
    \centering
    \includegraphics[width=0.35\textwidth]{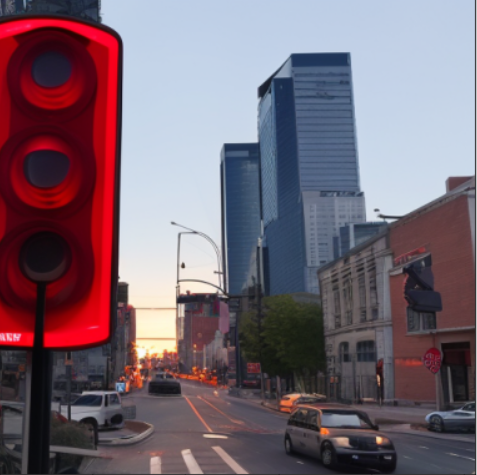}
    \caption{An image sample generated by Stable Diffusion 2 to extract local explanations.}
    \label{fig:local}
\end{figure}

\paragraph{Local explanations} provide edit paths based on the \textbf{I}, \textbf{D}, \textbf{R} operations realized for a specific generated image. For this reason, we employ a scene depicted in Figure \ref{fig:local}. 

According to YOLO-v8 with the default threshold $T_d$=0.6, the generated concepts are $S$=\{'car', 'car', 'traffic light', 'car', 'stop sign'\}, and ground truth concepts are $T$=\{'light', 'buildings'\}. The edit operations of total minimum cost 59.00 for this $S\rightarrow T$ transformation are:

\textbf{I}: \{ \}

\textbf{D}: \{'car', 'car', 'car'\}

\textbf{R}: \{'traffic light'$\rightarrow$'light', 'stop sign'$\rightarrow$'buildings'\}

When using YOLOS, the generated concepts are $S$=\{'car', 'traffic light', 'car', 'stop sign', 'traffic light', 'car', 'traffic light', 'traffic light', 'traffic light', 'traffic light', 'traffic light', 'traffic light', 'car', 'traffic light', 'traffic light', 'traffic light', 'traffic light', 'traffic light', 'car', 'traffic light', 'traffic light', 'traffic light', 'traffic light', 'car', 'car', 'traffic light', 'traffic light'\}, and the ground truth ones are $T$=\{'light', 'buildings'\}. By visually inspecting the image, YOLOS clearly overestimates the actual objects present, inducing noise in the generated concept set $S$. Nevertheless, our evaluation strategy successfully captures this overestimation, by suggesting the deletion of multiple concepts. Specifically, we obtain the following transformations of total cost 104.04:

\textbf{I}: \{ \}

\textbf{D}: \{'car', 'traffic light', 'car', 'traffic light', 'car', 'traffic light', 'traffic light', 'traffic light', 'traffic light', 'traffic light', 'traffic light', 'car', 'traffic light', 'traffic light', 'traffic light', 'traffic light', 'car', 'traffic light', 'traffic light', 'traffic light', 'traffic light', 'car', 'car', 'traffic light', 'traffic light'\}

\textbf{R}: \{'stop sign'$\rightarrow$'light', 'traffic light'$\rightarrow$'buildings'\}

\begin{table}[htp!]
\caption{Global explanations (\textbf{I} and \textbf{D} edits) for YOLO-v8 extracted concepts.}
\label{tab:global-id}
\begin{tabular}{c|>
{\centering\arraybackslash}p{6em}|>{\centering\arraybackslash}p{2em}>{\centering\arraybackslash}p{2.7em}>{\centering\arraybackslash}p{4em}>{\centering\arraybackslash}p{3em}>{\centering\arraybackslash}p{2.8em}>{\centering\arraybackslash}p{4.5em}}
\toprule
$T_d$ & $M$ & \textbf{I} & Freq \textbf{I} & \textbf{I} support & \textbf{D} & Freq \textbf{D} & \textbf{D} support \\
\midrule
\multirow{12}{1em}{0.5} & \multirow{3}{*}{stable diffusion} & street & 264 & 1.57\% & person & 2075 & 36.69\%  \\
& & table & 250 & 1.49\% & sheep & 363 & 6.42\% \\
& & tennis & 247 & 1.47\% & car & 252 & 4.46\% \\\cline{2-8}
& \multirow{3}{6em}{stable diffusion 2} & tennis & 253 & 1.57\% & person & 2177 & 34.55\%  \\
& & street & 242 & 1.51\% & sheep & 466 & 7.40\% \\
& & table & 237 & 1.48\% & car & 313 & 4.97\%  \\\cline{2-8}
 & \multirow{3}{6em}{protogen base} & tennis & 247 & 1.52\% & person & 2281 & 38.37\% \\
 & & street & 244 & 1.51\% & sheep & 317 & 5.33\%  \\
 &  & table & 229 & 1.41\% & car & 311 & 5.23\% \\\cline{2-8}
& \multirow{3}{6em}{protogen 5.8} & table & 270 & 1.52\% & person & 1564 & 33.26\%  \\
& & tennis & 265 & 1.50\% & car & 261 & 5.55\% \\
& & street & 241 & 1.36\% & umbrella & 251 & 5.34\% \\
\midrule
\multirow{12}{1em}{0.6}  
& \multirow{3}{*}{stable diffusion} & street & 290 & 1.58\% & person & 1572 & 38.89\%  \\
 & & table & 281 & 1.53\% & sheep & 311 & 7.69\% \\
 &  & tennis & 259 & 1.41\% & car & 158 & 3.91\%  \\\cline{2-8}
& \multirow{3}{6em}{stable diffusion 2} & table & 274 & 1.54\% & person & 1656 & 36.69\%  \\
 & & street & 269 & 1.51\% & sheep & 376 & 8.33\% \\
 & & tennis & 264 & 1.48\% & car & 203 & 4.50\% \\\cline{2-8}
 & \multirow{3}{6em}{protogen base} & street & 268 & 1.51\% & person & 1717 & 40.21\%  \\
 &  & table & 261 & 1.47\% & sheep & 254 & 5.95\% \\
 &  & tennis & 255 & 1.43\% & car & 197 & 4.61\%  \\\cline{2-8}
& \multirow{3}{*}{protogen 5.8}  & table & 303 & 1.58\% & person & 1220 & 35.68\%  \\
 & & tennis & 278 & 1.45\% & sheep & 198 & 5.79\%  \\
 & & street & 274 & 1.43\% & umbrella & 176 & 5.15\% \\
\midrule
\multirow{12}{1em}{0.7} 
 & \multirow{3}{6em}{stable diffusion } & table & 322 & 1.58\% & person & 1075 & 40.10\%  \\
 &  & street & 316 & 1.55\% & sheep & 254 & 9.47\% \\
 & & tennis & 268 & 1.31\% & donut & 122 & 4.55\% \\\cline{2-8}
& \multirow{3}{6em}{stable diffusion 2} & table & 313 & 1.58\% & person & 1134 & 38.60\% \\ 
& & street & 301 & 1.52\% & sheep & 291 & 9.90\%  \\
&  & tennis & 267 & 1.35\% & donut & 111 & 3.78\% \\\cline{2-8}
& \multirow{3}{*}{protogen base} & street & 300 & 1.52\% & person & 1189 & 42.16\%  \\
& & table & 289 & 1.46\% & sheep & 188 & 6.67\%  \\
&  & tennis & 262 & 1.32\% & umbrella & 143 & 5.07\%  \\\cline{2-8}

& \multirow{3}{*}{protogen 5.8} & table & 330 & 1.58\% & person & 884 & 38.30\%  \\
&& street & 299 & 1.43\% & sheep & 152 & 6.59\%  \\
&  & tennis & 287 & 1.37\% & umbrella & 130 & 5.63\%\\
\bottomrule
\end{tabular}
\end{table}

\paragraph{Global explanations} for all images are presented in Table \ref{tab:global-id} for \textbf{I}, \textbf{D} edits and Table \ref{tab:global-id2} for \textbf{R} edits. Results only involve YOLO-v8 extracted concepts, as YOLOS results in an overwhelming number of detected instances.
Top-3 results are demonstrated, i.e. the 3 most frequent insertion, deletions and replacements. \textbf{I} and \textbf{D} refers to concepts inserted or deleted respectively, while Freq \textbf{I}, \textbf{D} denotes how many times a specific concepts was inserted or deleted within all images. \textbf{I}, \textbf{D} support indicates the frequency a specific edit happens among all \textbf{I}, \textbf{D} edits respectively. As for \textbf{R}, support denotes the frequency of a transformation rule among all produced rules.

We can observe an obvious agreement between models; \textbf{I} edits include 'street', 'tennis' and 'table' concepts. It seems that the selected $M$ cannot efficiently generate the \textbf{I} concepts, or generated concepts are of low visual quality, so that their detection is not feasible with $T_d$=0.5, 0.6, 0.7. 
\textbf{D} edits mainly contain 'person', 'sheep', 'car', 'umbrella', 'donut' concepts, indicating some bias towards generating spurious instances of those concept categories. Finally, \textbf{R} edits refer to transforming 'person' to 'people', 'man' or 'woman'. Since 'person' is a YOLO category incorporating both genders, such transformations are somehow expected.

\begin{table}[htp!]
\caption{Global explanations (\textbf{R} edits) for YOLO-v8 extracted concepts.}
\label{tab:global-id2}
\begin{tabular}{c|>
{\centering\arraybackslash}p{3.2em}|>{\centering\arraybackslash}p{7.2em}>{\centering\arraybackslash}p{1.9em}>{\centering\arraybackslash}p{3.1em}|>
{\centering\arraybackslash}p{3.2em}|>{\centering\arraybackslash}p{7.2em}>{\centering\arraybackslash}p{1.9em}>{\centering\arraybackslash}p{3em}}
\toprule
$T_d$ & $M$ & \textbf{R} & Freq \textbf{R} & \textbf{R} support & $M$ & \textbf{R} & Freq \textbf{R} & \textbf{R} support \\
\midrule
\multirow{6}{*}{0.5} &  \multirow{3}{5em}{stable diffusion} & person $\rightarrow$ man & 1090 & 7.61\% & stable & person $\rightarrow$ man & 1115 & 7.51\%\\
& & person $\rightarrow$ people & 520 & 3.63\% &diffusion& person $\rightarrow$ people & 551 & 3.71\%\\
& & person $\rightarrow$ woman & 499 & 3.48\% & 2 & person $\rightarrow$ woman & 511 & 3.44\%\\
\cline{2-9}
& \multirow{2}{8em}{protogen} & person $\rightarrow$ man & 1101 & 7.47\% & \multirow{2}{5em}{protogen }  & person $\rightarrow$ man & 1061 & 7.91\% \\
&& person $\rightarrow$ people & 507 & 3.44\%&  & person $\rightarrow$ woman & 476 & 3.55\% \\
&base& person $\rightarrow$ woman & 500 & 3.39\% & 5.8 & person $\rightarrow$ people & 441 & 3.29\% \\
\midrule
\multirow{6}{*}{0.6} &  \multirow{3}{5em}{stable diffusion} & person $\rightarrow$ man & 1065 & 8.22\% & stable & person $\rightarrow$ man & 1087 & 8.11\%\\
& & person $\rightarrow$ people & 503 & 3.88\% & diffusion & person $\rightarrow$people & 536 & 4.00\%\\
& & person $\rightarrow$ woman & 481 & 3.71\% & 2 & person $\rightarrow$ woman & 482 & 3.60\%\\\cline{2-9}
& \multirow{2}{8em}{protogen} & person $\rightarrow$ man & 1080 & 8.04\% & \multirow{2}{5em}{protogen } &  person $\rightarrow$ man & 1035 & 8.53\% \\
& & person $\rightarrow$ people & 494 & 3.68\% & &person $\rightarrow$ woman & 449 & 3.70\% \\
& base & person $\rightarrow$ woman & 485 & 3.61\% & 5.8 &  person $\rightarrow$ people & 431 & 3.55\%\\
\midrule
\multirow{6}{*}{0.7} &  \multirow{3}{5em}{stable diffusion} & person $\rightarrow$ man & 1022 & 9.01\% & stable & person $\rightarrow$ man & 1033 & 8.79\%\\
& & person $\rightarrow$ people & 473 & 4.17\% & diffusion & person $\rightarrow$ people & 508 & 4.32\%\\
& & person $\rightarrow$ woman & 458 & 4.04\% & 2 & person $\rightarrow$ woman & 441 & 3.75\%\\\cline{2-9}
& \multirow{2}{8em}{protogen} & person $\rightarrow$ man & 1054 & 8.94\% & \multirow{2}{5em}{protogen } & person $\rightarrow$ man & 989 & 9.22\% \\
& & person $\rightarrow$ woman & 461 & 3.91\% && person $\rightarrow$ woman & 419 & 3.91\% \\
& base &person $\rightarrow$ people & 446 & 3.78\% & 5.8 & person $\rightarrow$ people & 408 & 3.80\% \\
\bottomrule
\end{tabular}
\end{table}

\section{Conclusion}
Conceptual approaches in generative evaluation is an underexplored field, which can provide some novel insights  regarding model quality and explainability of results. In our work, we propose a knowledge-driven explainable evaluation framework that suggests which concepts should be added, removed, or replaced for a generated image to approach its conditioning. Results on competitive tasks such as Story Visualization and Scene Generation illustrate the merits of such an approach, highlighting concepts that models cannot generate, or model biases towards generating excessive numbers of specific concept categories. As future work, we plan to expand our approach to other models and tasks and also incorporate alternative knowledge sources to examine how the produced edit paths conceptually deviate from the current ones.

\begin{acknowledgments}
The research work was supported by the Hellenic Foundation for Research and Innovation (HFRI) under the 3rd Call for HFRI PhD Fellowships (Fellowship Number 5537).
\end{acknowledgments}

\bibliography{sample-ceur}

\begin{thebibliography}{74}
\expandafter\ifx\csname natexlab\endcsname\relax\def\natexlab#1{#1}\fi
\providecommand{\url}[1]{\texttt{#1}}
\providecommand{\href}[2]{#2}
\providecommand{\path}[1]{#1}
\providecommand{\DOIprefix}{doi:}
\providecommand{\ArXivprefix}{arXiv:}
\providecommand{\URLprefix}{URL: }
\providecommand{\Pubmedprefix}{pmid:}
\providecommand{\doi}[1]{\href{http://dx.doi.org/#1}{\path{#1}}}
\providecommand{\Pubmed}[1]{\href{pmid:#1}{\path{#1}}}
\providecommand{\bibinfo}[2]{#2}
\ifx\xfnm\relax \def\xfnm[#1]{\unskip,\space#1}\fi
\bibitem[{Ho et~al.(2020)Ho, Jain, and Abbeel}]{diff}
\bibinfo{author}{J.~Ho}, \bibinfo{author}{A.~Jain},
  \bibinfo{author}{P.~Abbeel}, \bibinfo{title}{Denoising diffusion
  probabilistic models}, \bibinfo{year}{2020}. \URLprefix
  \url{https://arxiv.org/abs/2006.11239}.
  \DOIprefix\doi{10.48550/ARXIV.2006.11239}.
\bibitem[{Rombach et~al.(2022)Rombach, Blattmann, Lorenz, Esser, and
  Ommer}]{diffusion}
\bibinfo{author}{R.~Rombach}, \bibinfo{author}{A.~Blattmann},
  \bibinfo{author}{D.~Lorenz}, \bibinfo{author}{P.~Esser},
  \bibinfo{author}{B.~Ommer},
\newblock \bibinfo{title}{High-resolution image synthesis with latent diffusion
  models},
\newblock in: \bibinfo{booktitle}{Proceedings of the IEEE/CVF Conference on
  Computer Vision and Pattern Recognition (CVPR)}, \bibinfo{year}{2022}, pp.
  \bibinfo{pages}{10684--10695}.
\bibitem[{Ramesh et~al.(2021)Ramesh, Pavlov, Goh, Gray, Voss, Radford, Chen,
  and Sutskever}]{dalle}
\bibinfo{author}{A.~Ramesh}, \bibinfo{author}{M.~Pavlov},
  \bibinfo{author}{G.~Goh}, \bibinfo{author}{S.~Gray},
  \bibinfo{author}{C.~Voss}, \bibinfo{author}{A.~Radford},
  \bibinfo{author}{M.~Chen}, \bibinfo{author}{I.~Sutskever},
  \bibinfo{title}{Zero-shot text-to-image generation}, \bibinfo{year}{2021}.
  \href{http://arxiv.org/abs/2102.12092}{{\tt arXiv:2102.12092}}.
\bibitem[{Ramesh et~al.(2022)Ramesh, Dhariwal, Nichol, Chu, and Chen}]{dalle2}
\bibinfo{author}{A.~Ramesh}, \bibinfo{author}{P.~Dhariwal},
  \bibinfo{author}{A.~Nichol}, \bibinfo{author}{C.~Chu},
  \bibinfo{author}{M.~Chen}, \bibinfo{title}{Hierarchical text-conditional
  image generation with clip latents}, \bibinfo{year}{2022}. \URLprefix
  \url{https://arxiv.org/abs/2204.06125}.
  \DOIprefix\doi{10.48550/ARXIV.2204.06125}.
\bibitem[{Saharia et~al.(2022)Saharia, Chan, Saxena, Li, Whang, Denton,
  Ghasemipour, Ayan, Mahdavi, Lopes, Salimans, Ho, Fleet, and Norouzi}]{imagen}
\bibinfo{author}{C.~Saharia}, \bibinfo{author}{W.~Chan},
  \bibinfo{author}{S.~Saxena}, \bibinfo{author}{L.~Li},
  \bibinfo{author}{J.~Whang}, \bibinfo{author}{E.~Denton},
  \bibinfo{author}{S.~K.~S. Ghasemipour}, \bibinfo{author}{B.~K. Ayan},
  \bibinfo{author}{S.~S. Mahdavi}, \bibinfo{author}{R.~G. Lopes},
  \bibinfo{author}{T.~Salimans}, \bibinfo{author}{J.~Ho},
  \bibinfo{author}{D.~J. Fleet}, \bibinfo{author}{M.~Norouzi},
  \bibinfo{title}{Photorealistic text-to-image diffusion models with deep
  language understanding}, \bibinfo{year}{2022}.
  \href{http://arxiv.org/abs/2205.11487}{{\tt arXiv:2205.11487}}.
\bibitem[{Blattmann et~al.(2022)Blattmann, Rombach, Oktay, Müller, and
  Ommer}]{semi}
\bibinfo{author}{A.~Blattmann}, \bibinfo{author}{R.~Rombach},
  \bibinfo{author}{K.~Oktay}, \bibinfo{author}{J.~Müller},
  \bibinfo{author}{B.~Ommer}, \bibinfo{title}{Semi-parametric neural image
  synthesis}, \bibinfo{year}{2022}. \URLprefix
  \url{https://arxiv.org/abs/2204.11824}.
  \DOIprefix\doi{10.48550/ARXIV.2204.11824}.
\bibitem[{Kim and Lee(2023)}]{cc}
\bibinfo{author}{J.~Kim}, \bibinfo{author}{M.~Lee},
  \bibinfo{title}{Class-continuous conditional generative neural radiance
  field}, \bibinfo{year}{2023}. \URLprefix
  \url{https://arxiv.org/abs/2301.00950}.
  \DOIprefix\doi{10.48550/ARXIV.2301.00950}.
\bibitem[{Goodfellow et~al.(2014)Goodfellow, Pouget-Abadie, Mirza, Xu,
  Warde-Farley, Ozair, Courville, and Bengio}]{goodfellow2014gans}
\bibinfo{author}{I.~Goodfellow}, \bibinfo{author}{J.~Pouget-Abadie},
  \bibinfo{author}{M.~Mirza}, \bibinfo{author}{B.~Xu},
  \bibinfo{author}{D.~Warde-Farley}, \bibinfo{author}{S.~Ozair},
  \bibinfo{author}{A.~Courville}, \bibinfo{author}{Y.~Bengio},
\newblock \bibinfo{title}{Generative adversarial nets},
\newblock in: \bibinfo{editor}{Z.~Ghahramani}, \bibinfo{editor}{M.~Welling},
  \bibinfo{editor}{C.~Cortes}, \bibinfo{editor}{N.~Lawrence},
  \bibinfo{editor}{K.~Q. Weinberger} (Eds.), \bibinfo{booktitle}{Advances in
  Neural Information Processing Systems}, volume~\bibinfo{volume}{27},
  \bibinfo{publisher}{Curran Associates, Inc.}, \bibinfo{year}{2014}.
  \URLprefix
  \url{https://proceedings.neurips.cc/paper/2014/file/5ca3e9b122f61f8f06494c97b1afccf3-Paper.pdf}.
\bibitem[{Salimans et~al.(2016)Salimans, Goodfellow, Zaremba, Cheung, Radford,
  Chen, and Chen}]{is}
\bibinfo{author}{T.~Salimans}, \bibinfo{author}{I.~Goodfellow},
  \bibinfo{author}{W.~Zaremba}, \bibinfo{author}{V.~Cheung},
  \bibinfo{author}{A.~Radford}, \bibinfo{author}{X.~Chen},
  \bibinfo{author}{X.~Chen},
\newblock \bibinfo{title}{Improved techniques for training gans},
\newblock in: \bibinfo{editor}{D.~Lee}, \bibinfo{editor}{M.~Sugiyama},
  \bibinfo{editor}{U.~Luxburg}, \bibinfo{editor}{I.~Guyon},
  \bibinfo{editor}{R.~Garnett} (Eds.), \bibinfo{booktitle}{Advances in Neural
  Information Processing Systems}, volume~\bibinfo{volume}{29},
  \bibinfo{publisher}{Curran Associates, Inc.}, \bibinfo{year}{2016}.
  \URLprefix
  \url{https://proceedings.neurips.cc/paper/2016/file/8a3363abe792db2d8761d6403605aeb7-Paper.pdf}.
\bibitem[{Heusel et~al.(2017)Heusel, Ramsauer, Unterthiner, Nessler, and
  Hochreiter}]{fid}
\bibinfo{author}{M.~Heusel}, \bibinfo{author}{H.~Ramsauer},
  \bibinfo{author}{T.~Unterthiner}, \bibinfo{author}{B.~Nessler},
  \bibinfo{author}{S.~Hochreiter},
\newblock \bibinfo{title}{Gans trained by a two time-scale update rule converge
  to a local nash equilibrium},
\newblock in: \bibinfo{editor}{I.~Guyon}, \bibinfo{editor}{U.~V. Luxburg},
  \bibinfo{editor}{S.~Bengio}, \bibinfo{editor}{H.~Wallach},
  \bibinfo{editor}{R.~Fergus}, \bibinfo{editor}{S.~Vishwanathan},
  \bibinfo{editor}{R.~Garnett} (Eds.), \bibinfo{booktitle}{Advances in Neural
  Information Processing Systems}, volume~\bibinfo{volume}{30},
  \bibinfo{publisher}{Curran Associates, Inc.}, \bibinfo{year}{2017}.
  \URLprefix
  \url{https://proceedings.neurips.cc/paper/2017/file/8a1d694707eb0fefe65871369074926d-Paper.pdf}.
\bibitem[{Zhang et~al.(2018)Zhang, Isola, Efros, Shechtman, and Wang}]{lpips}
\bibinfo{author}{R.~Zhang}, \bibinfo{author}{P.~Isola}, \bibinfo{author}{A.~A.
  Efros}, \bibinfo{author}{E.~Shechtman}, \bibinfo{author}{O.~Wang},
\newblock \bibinfo{title}{The unreasonable effectiveness of deep features as a
  perceptual metric},
\newblock in: \bibinfo{booktitle}{CVPR}, \bibinfo{year}{2018}.
\bibitem[{Parmar et~al.(2022)Parmar, Zhang, and Zhu}]{cleanfid}
\bibinfo{author}{G.~Parmar}, \bibinfo{author}{R.~Zhang}, \bibinfo{author}{J.-Y.
  Zhu},
\newblock \bibinfo{title}{On aliased resizing and surprising subtleties in gan
  evaluation},
\newblock in: \bibinfo{booktitle}{CVPR}, \bibinfo{year}{2022}.
\bibitem[{Borji(2022)}]{eval}
\bibinfo{author}{A.~Borji},
\newblock \bibinfo{title}{Pros and cons of gan evaluation measures: New
  developments},
\newblock \bibinfo{journal}{Computer Vision and Image Understanding}
  \bibinfo{volume}{215} (\bibinfo{year}{2022}) \bibinfo{pages}{103329}.
  \URLprefix
  \url{https://www.sciencedirect.com/science/article/pii/S1077314221001685}.
  \DOIprefix\doi{https://doi.org/10.1016/j.cviu.2021.103329}.
\bibitem[{Soloveitchik et~al.(2021)Soloveitchik, Diskin, Morin, and
  Wiesel}]{c-fid}
\bibinfo{author}{M.~Soloveitchik}, \bibinfo{author}{T.~Diskin},
  \bibinfo{author}{E.~Morin}, \bibinfo{author}{A.~Wiesel},
  \bibinfo{title}{Conditional frechet inception distance},
  \bibinfo{year}{2021}. \URLprefix \url{https://arxiv.org/abs/2103.11521}.
  \DOIprefix\doi{10.48550/ARXIV.2103.11521}.
\bibitem[{Benny et~al.(2021)Benny, Galanti, Benaim, and Wolf}]{c-eval}
\bibinfo{author}{Y.~Benny}, \bibinfo{author}{T.~Galanti},
  \bibinfo{author}{S.~Benaim}, \bibinfo{author}{L.~Wolf},
\newblock \bibinfo{title}{Evaluation metrics for conditional image generation},
\newblock \bibinfo{journal}{International Journal of Computer Vision}
  \bibinfo{volume}{129} (\bibinfo{year}{2021}) \bibinfo{pages}{1712--1731}.
  \URLprefix \url{https://doi.org/10.1007%2Fs11263-020-01424-w}.
  \DOIprefix\doi{10.1007/s11263-020-01424-w}.
\bibitem[{Abhishek and Kamath(2022)}]{xaivision}
\bibinfo{author}{K.~Abhishek}, \bibinfo{author}{D.~Kamath},
  \bibinfo{title}{Attribution-based xai methods in computer vision: A review},
  \bibinfo{year}{2022}. \URLprefix \url{https://arxiv.org/abs/2211.14736}.
  \DOIprefix\doi{10.48550/ARXIV.2211.14736}.
\bibitem[{Buhrmester et~al.(2021)Buhrmester, MÃ¼nch, and Arens}]{xaivision2}
\bibinfo{author}{V.~Buhrmester}, \bibinfo{author}{D.~MÃ¼nch},
  \bibinfo{author}{M.~Arens},
\newblock \bibinfo{title}{Analysis of explainers of black box deep neural
  networks for computer vision: A survey},
\newblock \bibinfo{journal}{Machine Learning and Knowledge Extraction}
  \bibinfo{volume}{3} (\bibinfo{year}{2021}) \bibinfo{pages}{966--989}.
  \URLprefix \url{https://www.mdpi.com/2504-4990/3/4/48}.
  \DOIprefix\doi{10.3390/make3040048}.
\bibitem[{Nagisetty et~al.(2020)Nagisetty, Graves, Scott, and Ganesh}]{xaigan}
\bibinfo{author}{V.~Nagisetty}, \bibinfo{author}{L.~Graves},
  \bibinfo{author}{J.~Scott}, \bibinfo{author}{V.~Ganesh},
  \bibinfo{title}{xai-gan: Enhancing generative adversarial networks via
  explainable ai systems}, \bibinfo{year}{2020}. \URLprefix
  \url{https://arxiv.org/abs/2002.10438}.
  \DOIprefix\doi{10.48550/ARXIV.2002.10438}.
\bibitem[{Genovese et~al.(2019)Genovese, Piuri, and Scotti}]{xaigan2}
\bibinfo{author}{A.~Genovese}, \bibinfo{author}{V.~Piuri},
  \bibinfo{author}{F.~Scotti},
\newblock \bibinfo{title}{Towards explainable face aging with generative
  adversarial networks},
\newblock in: \bibinfo{booktitle}{2019 IEEE International Conference on Image
  Processing (ICIP)}, \bibinfo{year}{2019}, pp. \bibinfo{pages}{3806--3810}.
  \DOIprefix\doi{10.1109/ICIP.2019.8803616}.
\bibitem[{Kim and Park(2022)}]{xaigan3}
\bibinfo{author}{J.~Kim}, \bibinfo{author}{H.~Park},
\newblock \bibinfo{title}{Limited discriminator gan using explainable ai model
  for overfitting problem},
\newblock \bibinfo{journal}{ICT Express}  (\bibinfo{year}{2022}). \URLprefix
  \url{https://www.sciencedirect.com/science/article/pii/S240595952100179X}.
  \DOIprefix\doi{https://doi.org/10.1016/j.icte.2021.12.014}.
\bibitem[{Leiter et~al.(2022)Leiter, Lertvittayakumjorn, Fomicheva, Zhao, Gao,
  and Eger}]{nlpxai}
\bibinfo{author}{C.~Leiter}, \bibinfo{author}{P.~Lertvittayakumjorn},
  \bibinfo{author}{M.~Fomicheva}, \bibinfo{author}{W.~Zhao},
  \bibinfo{author}{Y.~Gao}, \bibinfo{author}{S.~Eger}, \bibinfo{title}{Towards
  explainable evaluation metrics for natural language generation},
  \bibinfo{year}{2022}. \URLprefix \url{https://arxiv.org/abs/2203.11131}.
  \DOIprefix\doi{10.48550/ARXIV.2203.11131}.
\bibitem[{Opitz and Frank(2021)}]{nlpxai2}
\bibinfo{author}{J.~Opitz}, \bibinfo{author}{A.~Frank},
\newblock \bibinfo{title}{Towards a decomposable metric for explainable
  evaluation of text generation from {AMR}},
\newblock in: \bibinfo{booktitle}{Proceedings of the 16th Conference of the
  European Chapter of the Association for Computational Linguistics: Main
  Volume}, \bibinfo{publisher}{Association for Computational Linguistics},
  \bibinfo{address}{Online}, \bibinfo{year}{2021}, pp.
  \bibinfo{pages}{1504--1518}. \URLprefix
  \url{https://aclanthology.org/2021.eacl-main.129}.
  \DOIprefix\doi{10.18653/v1/2021.eacl-main.129}.
\bibitem[{Lymperaiou et~al.(2022)Lymperaiou, Manoliadis,
  Menis~Mastromichalakis, Dervakos, and Stamou}]{nlpxai3}
\bibinfo{author}{M.~Lymperaiou}, \bibinfo{author}{G.~Manoliadis},
  \bibinfo{author}{O.~Menis~Mastromichalakis}, \bibinfo{author}{E.~G.
  Dervakos}, \bibinfo{author}{G.~Stamou},
\newblock \bibinfo{title}{Towards explainable evaluation of language models on
  the semantic similarity of visual concepts},
\newblock in: \bibinfo{booktitle}{Proceedings of the 29th International
  Conference on Computational Linguistics}, \bibinfo{publisher}{International
  Committee on Computational Linguistics}, \bibinfo{address}{Gyeongju, Republic
  of Korea}, \bibinfo{year}{2022}, pp. \bibinfo{pages}{3639--3658}. \URLprefix
  \url{https://aclanthology.org/2022.coling-1.321}.
\bibitem[{Filandrianos et~al.(2022)Filandrianos, Thomas, Dervakos, and
  Stamou}]{cece}
\bibinfo{author}{G.~Filandrianos}, \bibinfo{author}{K.~Thomas},
  \bibinfo{author}{E.~Dervakos}, \bibinfo{author}{G.~Stamou},
\newblock \bibinfo{title}{Conceptual edits as counterfactual explanations},
\newblock in: \bibinfo{booktitle}{Proceedings of the AAAI 2022 Spring Symposium
  on Machine Learning and Knowledge Engineering for Hybrid Intelligence
  (AAAI-MAKE 2022), Stanford University, Palo Alto, California, USA},
  \bibinfo{year}{2022}.
\bibitem[{Mirza and Osindero(2014)}]{mirza2014condgan}
\bibinfo{author}{M.~Mirza}, \bibinfo{author}{S.~Osindero},
\newblock \bibinfo{title}{Conditional generative adversarial nets},
\newblock \bibinfo{journal}{CoRR} \bibinfo{volume}{abs/1411.1784}
  (\bibinfo{year}{2014}). \URLprefix \url{http://arxiv.org/abs/1411.1784}.
  \href{http://arxiv.org/abs/1411.1784}{{\tt arXiv:1411.1784}}.
\bibitem[{Odena et~al.(2017)Odena, Olah, and Shlens}]{odena2017conditional}
\bibinfo{author}{A.~Odena}, \bibinfo{author}{C.~Olah},
  \bibinfo{author}{J.~Shlens}, \bibinfo{title}{Conditional image synthesis with
  auxiliary classifier gans}, \bibinfo{year}{2017}.
  \href{http://arxiv.org/abs/1610.09585}{{\tt arXiv:1610.09585}}.
\bibitem[{Miyato and Koyama(2018)}]{miyato2018cgans}
\bibinfo{author}{T.~Miyato}, \bibinfo{author}{M.~Koyama}, \bibinfo{title}{cgans
  with projection discriminator}, \bibinfo{year}{2018}.
  \href{http://arxiv.org/abs/1802.05637}{{\tt arXiv:1802.05637}}.
\bibitem[{Zhang et~al.(2019)Zhang, Goodfellow, Metaxas, and
  Odena}]{zhang2019sagan}
\bibinfo{author}{H.~Zhang}, \bibinfo{author}{I.~Goodfellow},
  \bibinfo{author}{D.~Metaxas}, \bibinfo{author}{A.~Odena},
  \bibinfo{title}{Self-attention generative adversarial networks},
  \bibinfo{year}{2019}. \href{http://arxiv.org/abs/1805.08318}{{\tt
  arXiv:1805.08318}}.
\bibitem[{Miyato et~al.(2018)Miyato, Kataoka, Koyama, and
  Yoshida}]{miyato2018spectral}
\bibinfo{author}{T.~Miyato}, \bibinfo{author}{T.~Kataoka},
  \bibinfo{author}{M.~Koyama}, \bibinfo{author}{Y.~Yoshida},
  \bibinfo{title}{Spectral normalization for generative adversarial networks},
  \bibinfo{year}{2018}. \href{http://arxiv.org/abs/1802.05957}{{\tt
  arXiv:1802.05957}}.
\bibitem[{Heusel et~al.(2018)Heusel, Ramsauer, Unterthiner, Nessler, and
  Hochreiter}]{heusel2018ttur}
\bibinfo{author}{M.~Heusel}, \bibinfo{author}{H.~Ramsauer},
  \bibinfo{author}{T.~Unterthiner}, \bibinfo{author}{B.~Nessler},
  \bibinfo{author}{S.~Hochreiter}, \bibinfo{title}{Gans trained by a two
  time-scale update rule converge to a local nash equilibrium},
  \bibinfo{year}{2018}. \href{http://arxiv.org/abs/1706.08500}{{\tt
  arXiv:1706.08500}}.
\bibitem[{Rombach et~al.(2021)Rombach, Blattmann, Lorenz, Esser, and
  Ommer}]{diff0}
\bibinfo{author}{R.~Rombach}, \bibinfo{author}{A.~Blattmann},
  \bibinfo{author}{D.~Lorenz}, \bibinfo{author}{P.~Esser},
  \bibinfo{author}{B.~Ommer}, \bibinfo{title}{High-resolution image synthesis
  with latent diffusion models}, \bibinfo{year}{2021}.
  \href{http://arxiv.org/abs/2112.10752}{{\tt arXiv:2112.10752}}.
\bibitem[{Radford et~al.(2021)Radford, Kim, Hallacy, Ramesh, Goh, Agarwal,
  Sastry, Askell, Mishkin, Clark, Krueger, and Sutskever}]{clip}
\bibinfo{author}{A.~Radford}, \bibinfo{author}{J.~Kim},
  \bibinfo{author}{C.~Hallacy}, \bibinfo{author}{A.~Ramesh},
  \bibinfo{author}{G.~Goh}, \bibinfo{author}{S.~Agarwal},
  \bibinfo{author}{G.~Sastry}, \bibinfo{author}{A.~Askell},
  \bibinfo{author}{P.~Mishkin}, \bibinfo{author}{J.~Clark},
  \bibinfo{author}{G.~Krueger}, \bibinfo{author}{I.~Sutskever},
  \bibinfo{title}{Learning transferable visual models from natural language
  supervision}, \bibinfo{year}{2021}.
\bibitem[{Saharia et~al.(2022)Saharia, Chan, Saxena, Li, Whang, Denton,
  Ghasemipour, Ayan, Mahdavi, Lopes, Salimans, Ho, Fleet, and Norouzi}]{diff1}
\bibinfo{author}{C.~Saharia}, \bibinfo{author}{W.~Chan},
  \bibinfo{author}{S.~Saxena}, \bibinfo{author}{L.~Li},
  \bibinfo{author}{J.~Whang}, \bibinfo{author}{E.~Denton},
  \bibinfo{author}{S.~K.~S. Ghasemipour}, \bibinfo{author}{B.~K. Ayan},
  \bibinfo{author}{S.~S. Mahdavi}, \bibinfo{author}{R.~G. Lopes},
  \bibinfo{author}{T.~Salimans}, \bibinfo{author}{J.~Ho},
  \bibinfo{author}{D.~J. Fleet}, \bibinfo{author}{M.~Norouzi},
  \bibinfo{title}{Photorealistic text-to-image diffusion models with deep
  language understanding}, \bibinfo{year}{2022}. \URLprefix
  \url{https://arxiv.org/abs/2205.11487}.
  \DOIprefix\doi{10.48550/ARXIV.2205.11487}.
\bibitem[{Raffel et~al.(2020)Raffel, Shazeer, Roberts, Lee, Narang, Matena,
  Zhou, Li, and Liu}]{t5}
\bibinfo{author}{C.~Raffel}, \bibinfo{author}{N.~Shazeer},
  \bibinfo{author}{A.~Roberts}, \bibinfo{author}{K.~Lee},
  \bibinfo{author}{S.~Narang}, \bibinfo{author}{M.~Matena},
  \bibinfo{author}{Y.~Zhou}, \bibinfo{author}{W.~Li}, \bibinfo{author}{P.~J.
  Liu},
\newblock \bibinfo{title}{Exploring the limits of transfer learning with a
  unified text-to-text transformer},
\newblock \bibinfo{journal}{Journal of Machine Learning Research}
  \bibinfo{volume}{21} (\bibinfo{year}{2020}) \bibinfo{pages}{1--67}.
  \URLprefix \url{http://jmlr.org/papers/v21/20-074.html}.
\bibitem[{Ruiz et~al.(2022)Ruiz, Li, Jampani, Pritch, Rubinstein, and
  Aberman}]{diff2}
\bibinfo{author}{N.~Ruiz}, \bibinfo{author}{Y.~Li},
  \bibinfo{author}{V.~Jampani}, \bibinfo{author}{Y.~Pritch},
  \bibinfo{author}{M.~Rubinstein}, \bibinfo{author}{K.~Aberman},
  \bibinfo{title}{Dreambooth: Fine tuning text-to-image diffusion models for
  subject-driven generation}, \bibinfo{year}{2022}. \URLprefix
  \url{https://arxiv.org/abs/2208.12242}.
  \DOIprefix\doi{10.48550/ARXIV.2208.12242}.
\bibitem[{Reed et~al.(2016{\natexlab{a}})Reed, Akata, Yan, Logeswaran, Schiele,
  and Lee}]{reed2016t2i}
\bibinfo{author}{S.~E. Reed}, \bibinfo{author}{Z.~Akata},
  \bibinfo{author}{X.~Yan}, \bibinfo{author}{L.~Logeswaran},
  \bibinfo{author}{B.~Schiele}, \bibinfo{author}{H.~Lee},
\newblock \bibinfo{title}{Generative adversarial text to image synthesis},
\newblock \bibinfo{journal}{CoRR} \bibinfo{volume}{abs/1605.05396}
  (\bibinfo{year}{2016}{\natexlab{a}}). \URLprefix
  \url{http://arxiv.org/abs/1605.05396}.
  \href{http://arxiv.org/abs/1605.05396}{{\tt arXiv:1605.05396}}.
\bibitem[{Reed et~al.(2016{\natexlab{b}})Reed, Akata, Mohan, Tenka, Schiele,
  and Lee}]{reed2016t2i2}
\bibinfo{author}{S.~E. Reed}, \bibinfo{author}{Z.~Akata},
  \bibinfo{author}{S.~Mohan}, \bibinfo{author}{S.~Tenka},
  \bibinfo{author}{B.~Schiele}, \bibinfo{author}{H.~Lee},
\newblock \bibinfo{title}{Learning what and where to draw},
\newblock \bibinfo{journal}{CoRR} \bibinfo{volume}{abs/1610.02454}
  (\bibinfo{year}{2016}{\natexlab{b}}). \URLprefix
  \url{http://arxiv.org/abs/1610.02454}.
  \href{http://arxiv.org/abs/1610.02454}{{\tt arXiv:1610.02454}}.
\bibitem[{Zhang et~al.(2017)Zhang, Xu, and Li}]{stackgan}
\bibinfo{author}{H.~Zhang}, \bibinfo{author}{T.~Xu}, \bibinfo{author}{H.~Li},
\newblock \bibinfo{title}{Stackgan: Text to photo-realistic image synthesis
  with stacked generative adversarial networks},
\newblock \bibinfo{year}{2017}, pp. \bibinfo{pages}{5908--5916}.
  \DOIprefix\doi{10.1109/ICCV.2017.629}.
\bibitem[{Zhang et~al.(2018)Zhang, Xu, Li, Zhang, Wang, Huang, and
  Metaxas}]{zhang2018stackgan}
\bibinfo{author}{H.~Zhang}, \bibinfo{author}{T.~Xu}, \bibinfo{author}{H.~Li},
  \bibinfo{author}{S.~Zhang}, \bibinfo{author}{X.~Wang},
  \bibinfo{author}{X.~Huang}, \bibinfo{author}{D.~Metaxas},
  \bibinfo{title}{Stackgan++: Realistic image synthesis with stacked generative
  adversarial networks}, \bibinfo{year}{2018}.
  \href{http://arxiv.org/abs/1710.10916}{{\tt arXiv:1710.10916}}.
\bibitem[{Xu et~al.(2018)Xu, Zhang, Huang, Zhang, Gan, Huang, and
  He}]{xu2018attngan}
\bibinfo{author}{T.~Xu}, \bibinfo{author}{P.~Zhang},
  \bibinfo{author}{Q.~Huang}, \bibinfo{author}{H.~Zhang},
  \bibinfo{author}{Z.~Gan}, \bibinfo{author}{X.~Huang},
  \bibinfo{author}{X.~He},
\newblock \bibinfo{title}{Attngan: Fine-grained text to image generation with
  attentional generative adversarial networks},
\newblock in: \bibinfo{booktitle}{CVPR 2018}, \bibinfo{year}{2018}.
\bibitem[{Tan et~al.(2019)Tan, Liu, Li, Zhang, and Yin}]{segan}
\bibinfo{author}{H.~Tan}, \bibinfo{author}{X.~Liu}, \bibinfo{author}{X.~Li},
  \bibinfo{author}{Y.~Zhang}, \bibinfo{author}{B.~Yin},
\newblock \bibinfo{title}{Semantics-enhanced adversarial nets for text-to-image
  synthesis},
\newblock in: \bibinfo{booktitle}{2019 IEEE/CVF International Conference on
  Computer Vision (ICCV)}, \bibinfo{year}{2019}, pp.
  \bibinfo{pages}{10500--10509}. \DOIprefix\doi{10.1109/ICCV.2019.01060}.
\bibitem[{Zhu et~al.(2019)Zhu, Pan, Chen, and Yang}]{zhu2019dmgan}
\bibinfo{author}{M.~Zhu}, \bibinfo{author}{P.~Pan}, \bibinfo{author}{W.~Chen},
  \bibinfo{author}{Y.~Yang},
\newblock \bibinfo{title}{{DM-GAN:} dynamic memory generative adversarial
  networks for text-to-image synthesis},
\newblock \bibinfo{journal}{CoRR} \bibinfo{volume}{abs/1904.01310}
  (\bibinfo{year}{2019}). \URLprefix \url{http://arxiv.org/abs/1904.01310}.
  \href{http://arxiv.org/abs/1904.01310}{{\tt arXiv:1904.01310}}.
\bibitem[{Li et~al.(2019)Li, Gan, Shen, Liu, Cheng, Wu, Carin, Carlson, and
  Gao}]{storygan}
\bibinfo{author}{Y.~Li}, \bibinfo{author}{Z.~Gan}, \bibinfo{author}{Y.~Shen},
  \bibinfo{author}{J.~Liu}, \bibinfo{author}{Y.~Cheng},
  \bibinfo{author}{Y.~Wu}, \bibinfo{author}{L.~Carin}, \bibinfo{author}{D.~E.
  Carlson}, \bibinfo{author}{J.~Gao},
\newblock \bibinfo{title}{Storygan: A sequential conditional gan for story
  visualization},
\newblock \bibinfo{journal}{2019 IEEE/CVF Conference on Computer Vision and
  Pattern Recognition (CVPR)}  (\bibinfo{year}{2019})
  \bibinfo{pages}{6322--6331}.
\bibitem[{Li et~al.(2020)Li, Kong, and Zhou}]{li2020storygan}
\bibinfo{author}{C.~Li}, \bibinfo{author}{L.~Kong}, \bibinfo{author}{Z.~Zhou},
\newblock \bibinfo{title}{Improved-storygan for sequential images
  visualization},
\newblock \bibinfo{journal}{Journal of Visual Communication and Image
  Representation} \bibinfo{volume}{73} (\bibinfo{year}{2020})
  \bibinfo{pages}{102956}. \URLprefix
  \url{https://www.sciencedirect.com/science/article/pii/S1047320320301826}.
  \DOIprefix\doi{https://doi.org/10.1016/j.jvcir.2020.102956}.
\bibitem[{Maharana et~al.(2021)Maharana, Hannan, and
  Bansal}]{Maharana2021ImprovingGA}
\bibinfo{author}{A.~Maharana}, \bibinfo{author}{D.~Hannan},
  \bibinfo{author}{M.~Bansal},
\newblock \bibinfo{title}{Improving generation and evaluation of visual stories
  via semantic consistency},
\newblock \bibinfo{journal}{ArXiv} \bibinfo{volume}{abs/2105.10026}
  (\bibinfo{year}{2021}).
\bibitem[{Maharana and Bansal(2021)}]{Maharana2021IntegratingVL}
\bibinfo{author}{A.~Maharana}, \bibinfo{author}{M.~Bansal},
\newblock \bibinfo{title}{Integrating visuospatial, linguistic, and commonsense
  structure into story visualization},
\newblock \bibinfo{journal}{ArXiv} \bibinfo{volume}{abs/2110.10834}
  (\bibinfo{year}{2021}).
\bibitem[{Tsakas et~al.(2023)Tsakas, Lymperaiou, Filandrianos, and
  Stamou}]{impartial}
\bibinfo{author}{N.~Tsakas}, \bibinfo{author}{M.~Lymperaiou},
  \bibinfo{author}{G.~Filandrianos}, \bibinfo{author}{G.~Stamou},
  \bibinfo{title}{An impartial transformer for story visualization},
  \bibinfo{year}{2023}. \URLprefix \url{https://arxiv.org/abs/2301.03563}.
  \DOIprefix\doi{10.48550/ARXIV.2301.03563}.
\bibitem[{Horé and Ziou(2010)}]{psnr}
\bibinfo{author}{A.~Horé}, \bibinfo{author}{D.~Ziou},
\newblock \bibinfo{title}{Image quality metrics: Psnr vs. ssim},
\newblock in: \bibinfo{booktitle}{2010 20th International Conference on Pattern
  Recognition}, \bibinfo{year}{2010}, pp. \bibinfo{pages}{2366--2369}.
  \DOIprefix\doi{10.1109/ICPR.2010.579}.
\bibitem[{Bau et~al.(2019)Bau, Zhu, Wulff, Peebles, Strobelt, Zhou, and
  Torralba}]{gan-cannot}
\bibinfo{author}{D.~Bau}, \bibinfo{author}{J.-Y. Zhu},
  \bibinfo{author}{J.~Wulff}, \bibinfo{author}{W.~Peebles},
  \bibinfo{author}{H.~Strobelt}, \bibinfo{author}{B.~Zhou},
  \bibinfo{author}{A.~Torralba}, \bibinfo{title}{Seeing what a gan cannot
  generate}, \bibinfo{year}{2019}. \URLprefix
  \url{https://arxiv.org/abs/1910.11626}.
  \DOIprefix\doi{10.48550/ARXIV.1910.11626}.
\bibitem[{Shen et~al.(2019)Shen, Gu, Tang, and Zhou}]{face}
\bibinfo{author}{Y.~Shen}, \bibinfo{author}{J.~Gu}, \bibinfo{author}{X.~Tang},
  \bibinfo{author}{B.~Zhou},
\newblock \bibinfo{title}{Interpreting the latent space of gans for semantic
  face editing},
\newblock \bibinfo{journal}{2020 IEEE/CVF Conference on Computer Vision and
  Pattern Recognition (CVPR)}  (\bibinfo{year}{2019})
  \bibinfo{pages}{9240--9249}.
\bibitem[{Shen et~al.(2020)Shen, Yang, Tang, and Zhou}]{interface}
\bibinfo{author}{Y.~Shen}, \bibinfo{author}{C.~Yang},
  \bibinfo{author}{X.~Tang}, \bibinfo{author}{B.~Zhou},
\newblock \bibinfo{title}{Interfacegan: Interpreting the disentangled face
  representation learned by gans},
\newblock \bibinfo{journal}{IEEE Transactions on Pattern Analysis and Machine
  Intelligence} \bibinfo{volume}{44} (\bibinfo{year}{2020})
  \bibinfo{pages}{2004--2018}.
\bibitem[{Chai et~al.(2021)Chai, Wulff, and Isola}]{compositionality}
\bibinfo{author}{L.~Chai}, \bibinfo{author}{J.~Wulff},
  \bibinfo{author}{P.~Isola},
\newblock \bibinfo{title}{Using latent space regression to analyze and leverage
  compositionality in gans},
\newblock \bibinfo{journal}{ArXiv} \bibinfo{volume}{abs/2103.10426}
  (\bibinfo{year}{2021}).
\bibitem[{Wen et~al.(2021)Wen, Benitez-Quiroz, Feng, and Martinez}]{diamond}
\bibinfo{author}{J.~Wen}, \bibinfo{author}{F.~Benitez-Quiroz},
  \bibinfo{author}{Q.~Feng}, \bibinfo{author}{A.~M. Martinez},
\newblock \bibinfo{title}{Diamond in the rough: Improving image realism by
  traversing the gan latent space},
\newblock \bibinfo{journal}{ArXiv} \bibinfo{volume}{abs/2104.05518}
  (\bibinfo{year}{2021}).
\bibitem[{Nauta et~al.(2022)Nauta, Trienes, Pathak, Nguyen, Peters, Schmitt,
  Schlötterer, van Keulen, and Seifert}]{xai}
\bibinfo{author}{M.~Nauta}, \bibinfo{author}{J.~Trienes},
  \bibinfo{author}{S.~Pathak}, \bibinfo{author}{E.~Nguyen},
  \bibinfo{author}{M.~Peters}, \bibinfo{author}{Y.~Schmitt},
  \bibinfo{author}{J.~Schlötterer}, \bibinfo{author}{M.~van Keulen},
  \bibinfo{author}{C.~Seifert}, \bibinfo{title}{From anecdotal evidence to
  quantitative evaluation methods: A systematic review on evaluating
  explainable ai}, \bibinfo{year}{2022}. \URLprefix
  \url{https://arxiv.org/abs/2201.08164}.
  \DOIprefix\doi{10.48550/ARXIV.2201.08164}.
\bibitem[{Zhao et~al.(2020)Zhao, Oyama, and Kurihara}]{ijcai2020}
\bibinfo{author}{W.~Zhao}, \bibinfo{author}{S.~Oyama},
  \bibinfo{author}{M.~Kurihara},
\newblock \bibinfo{title}{Generating natural counterfactual visual
  explanations},
\newblock in: \bibinfo{editor}{C.~Bessiere} (Ed.),
  \bibinfo{booktitle}{Proceedings of the Twenty-Ninth International Joint
  Conference on Artificial Intelligence, {IJCAI-20}},
  \bibinfo{publisher}{International Joint Conferences on Artificial
  Intelligence Organization}, \bibinfo{year}{2020}, pp.
  \bibinfo{pages}{5204--5205}. \URLprefix
  \url{https://doi.org/10.24963/ijcai.2020/742}.
  \DOIprefix\doi{10.24963/ijcai.2020/742}, \bibinfo{note}{doctoral Consortium}.
\bibitem[{Poyiadzi et~al.(2020)Poyiadzi, Sokol, Santos-Rodr{\'i}guez, Bie, and
  Flach}]{Poyiadzi2020FACEFA}
\bibinfo{author}{R.~Poyiadzi}, \bibinfo{author}{K.~Sokol},
  \bibinfo{author}{R.~Santos-Rodr{\'i}guez}, \bibinfo{author}{T.~D. Bie},
  \bibinfo{author}{P.~A. Flach},
\newblock \bibinfo{title}{Face: Feasible and actionable counterfactual
  explanations},
\newblock \bibinfo{journal}{Proceedings of the AAAI/ACM Conference on AI,
  Ethics, and Society}  (\bibinfo{year}{2020}).
\bibitem[{Goyal et~al.(2019)Goyal, Wu, Ernst, Batra, Parikh, and Lee}]{cve}
\bibinfo{author}{Y.~Goyal}, \bibinfo{author}{Z.~Wu},
  \bibinfo{author}{J.~Ernst}, \bibinfo{author}{D.~Batra},
  \bibinfo{author}{D.~Parikh}, \bibinfo{author}{S.~Lee},
\newblock \bibinfo{title}{Counterfactual visual explanations},
\newblock in: \bibinfo{editor}{K.~Chaudhuri},
  \bibinfo{editor}{R.~Salakhutdinov} (Eds.), \bibinfo{booktitle}{Proceedings of
  the 36th International Conference on Machine Learning},
  volume~\bibinfo{volume}{97} of \textit{\bibinfo{series}{Proceedings of
  Machine Learning Research}}, \bibinfo{publisher}{PMLR}, \bibinfo{year}{2019},
  pp. \bibinfo{pages}{2376--2384}. \URLprefix
  \url{https://proceedings.mlr.press/v97/goyal19a.html}.
\bibitem[{Fellbaum(1998)}]{wordnet}
\bibinfo{author}{C.~Fellbaum},
\newblock \bibinfo{title}{Wordnet: An electronic lexical database}
  (\bibinfo{year}{1998}).
\bibitem[{Browne and Swift(2020)}]{counterfactual1}
\bibinfo{author}{K.~Browne}, \bibinfo{author}{B.~Swift},
\newblock \bibinfo{title}{Semantics and explanation: why counterfactual
  explanations produce adversarial examples in deep neural networks},
\newblock \bibinfo{journal}{ArXiv} \bibinfo{volume}{abs/2012.10076}
  (\bibinfo{year}{2020}).
\bibitem[{Johnson et~al.(2017)Johnson, Hariharan, van~der Maaten, Fei-Fei,
  Zitnick, and Girshick}]{clevr}
\bibinfo{author}{J.~Johnson}, \bibinfo{author}{B.~Hariharan},
  \bibinfo{author}{L.~van~der Maaten}, \bibinfo{author}{L.~Fei-Fei},
  \bibinfo{author}{C.~L. Zitnick}, \bibinfo{author}{R.~B. Girshick},
\newblock \bibinfo{title}{Clevr: A diagnostic dataset for compositional
  language and elementary visual reasoning},
\newblock \bibinfo{journal}{2017 IEEE Conference on Computer Vision and Pattern
  Recognition (CVPR)}  (\bibinfo{year}{2017}) \bibinfo{pages}{1988--1997}.
\bibitem[{Lin et~al.(2015)Lin, Maire, Belongie, Bourdev, Girshick, Hays,
  Perona, Ramanan, Zitnick, and Dollár}]{coco}
\bibinfo{author}{T.-Y. Lin}, \bibinfo{author}{M.~Maire},
  \bibinfo{author}{S.~Belongie}, \bibinfo{author}{L.~Bourdev},
  \bibinfo{author}{R.~Girshick}, \bibinfo{author}{J.~Hays},
  \bibinfo{author}{P.~Perona}, \bibinfo{author}{D.~Ramanan},
  \bibinfo{author}{C.~L. Zitnick}, \bibinfo{author}{P.~Dollár},
  \bibinfo{title}{Microsoft coco: Common objects in context},
  \bibinfo{year}{2015}. \href{http://arxiv.org/abs/1405.0312}{{\tt
  arXiv:1405.0312}}.
\bibitem[{sta(face{\natexlab{a}})}]{stable1}
\bibinfo{title}{Stable diffusion v1.4},
  \bibinfo{year}{Huggingface}{\natexlab{a}}. \URLprefix
  \url{https://huggingface.co/CompVis/stable-diffusion-v1-4}.
\bibitem[{sta(face{\natexlab{b}})}]{stable2}
\bibinfo{title}{Stable diffusion 2 base},
  \bibinfo{year}{Huggingface}{\natexlab{b}}. \URLprefix
  \url{https://huggingface.co/stabilityai/stable-diffusion-2-base}.
\bibitem[{pro(face{\natexlab{a}})}]{protogen}
\bibinfo{title}{Protogen x3.4}, \bibinfo{year}{Huggingface}{\natexlab{a}}.
  \URLprefix
  \url{https://huggingface.co/darkstorm2150/Protogen_x3.4_Official_Release}.
\bibitem[{pro(face{\natexlab{b}})}]{protogen58}
\bibinfo{title}{Protogen x5.8}, \bibinfo{year}{Huggingface}{\natexlab{b}}.
  \URLprefix
  \url{https://huggingface.co/darkstorm2150/Protogen_x5.8_Official_Release}.
\bibitem[{Park et~al.(2019)Park, Liu, Wang, and Zhu}]{spade}
\bibinfo{author}{T.~Park}, \bibinfo{author}{M.-Y. Liu}, \bibinfo{author}{T.-C.
  Wang}, \bibinfo{author}{J.-Y. Zhu}, \bibinfo{title}{Semantic image synthesis
  with spatially-adaptive normalization}, \bibinfo{year}{2019}.
  \href{http://arxiv.org/abs/1903.07291}{{\tt arXiv:1903.07291}}.
\bibitem[{Zareian et~al.(2020)Zareian, Wang, You, and Chang}]{g2i19}
\bibinfo{author}{A.~Zareian}, \bibinfo{author}{Z.~Wang},
  \bibinfo{author}{H.~You}, \bibinfo{author}{S.-F. Chang},
  \bibinfo{title}{Learning visual commonsense for robust scene graph
  generation}, \bibinfo{year}{2020}.
  \href{http://arxiv.org/abs/2006.09623}{{\tt arXiv:2006.09623}}.
\bibitem[{Ma et~al.(2020)Ma, Zhao, and Sigal}]{g2i17}
\bibinfo{author}{K.~Ma}, \bibinfo{author}{B.~Zhao}, \bibinfo{author}{L.~Sigal},
  \bibinfo{title}{Attribute-guided image generation from layout},
  \bibinfo{year}{2020}. \href{http://arxiv.org/abs/2008.11932}{{\tt
  arXiv:2008.11932}}.
\bibitem[{Li et~al.(2020)Li, Cheng, Gan, Yu, Wang, and Liu}]{g2i16}
\bibinfo{author}{Y.~Li}, \bibinfo{author}{Y.~Cheng}, \bibinfo{author}{Z.~Gan},
  \bibinfo{author}{L.~Yu}, \bibinfo{author}{L.~Wang}, \bibinfo{author}{J.~Liu},
  \bibinfo{title}{Bachgan: High-resolution image synthesis from salient object
  layout}, \bibinfo{year}{2020}. \href{http://arxiv.org/abs/2003.11690}{{\tt
  arXiv:2003.11690}}.
\bibitem[{Sun and Wu(2021)}]{g2i15}
\bibinfo{author}{W.~Sun}, \bibinfo{author}{T.~Wu}, \bibinfo{title}{Learning
  layout and style reconfigurable gans for controllable image synthesis},
  \bibinfo{year}{2021}. \href{http://arxiv.org/abs/2003.11571}{{\tt
  arXiv:2003.11571}}.
\bibitem[{Jocher et~al.(2023)Jocher, Chaurasia, and Qiu}]{yolov8}
\bibinfo{author}{G.~Jocher}, \bibinfo{author}{A.~Chaurasia},
  \bibinfo{author}{J.~Qiu}, \bibinfo{title}{Yolo by ultralytics},
  \bibinfo{year}{2023}. \URLprefix
  \url{https://github.com/ultralytics/ultralytics}.
\bibitem[{Fang et~al.(2021)Fang, Liao, Wang, Fang, Qi, Wu, Niu, and
  Liu}]{yolos}
\bibinfo{author}{Y.~Fang}, \bibinfo{author}{B.~Liao},
  \bibinfo{author}{X.~Wang}, \bibinfo{author}{J.~Fang},
  \bibinfo{author}{J.~Qi}, \bibinfo{author}{R.~Wu}, \bibinfo{author}{J.~Niu},
  \bibinfo{author}{W.~Liu},
\newblock \bibinfo{title}{You only look at one sequence: Rethinking transformer
  in vision through object detection},
\newblock \bibinfo{journal}{CoRR} \bibinfo{volume}{abs/2106.00666}
  (\bibinfo{year}{2021}). \URLprefix \url{https://arxiv.org/abs/2106.00666}.
  \href{http://arxiv.org/abs/2106.00666}{{\tt arXiv:2106.00666}}.
\bibitem[{spa(paCy)}]{spacy}
\bibinfo{title}{Industrial-strength natural language processing},
  \bibinfo{year}{spaCy}. \URLprefix \url{https://spacy.io/}.
\bibitem[{Agrawal and Srikant(1994)}]{apriori}
\bibinfo{author}{R.~Agrawal}, \bibinfo{author}{R.~Srikant},
\newblock \bibinfo{title}{Fast algorithms for mining association rules},
\newblock in: \bibinfo{booktitle}{Proc. of 20th Intl. Conf. on VLDB},
  \bibinfo{year}{1994}, pp. \bibinfo{pages}{487--499}.

\end{thebibliography}

\appendix

\section{Online Resources}

The following models were used for image generation:
\begin{itemize}
\item
\href{https://huggingface.co/models?pipeline_tag=text-to-image&sort=downloads}{Huggingface text-to-image generation models}.
\item
\href{https://huggingface.co/CompVis/stable-diffusion-v1-4}{Stable diffusion v1.4}
\item
\href{https://huggingface.co/stabilityai/stable-diffusion-2}{Stable diffusion 2}
\item
\href{https://huggingface.co/darkstorm2150/Protogen_x3.4_Official_Release}{Protogen x3.4}
\item
\href{https://huggingface.co/darkstorm2150/Protogen_x5.8_Official_Release}{Protogen x5.8}
\end{itemize}

The following web-page was used for retrieving already generated images with Stable-Diffusion:
\begin{itemize}
    \item \href{https://lexica.art/}{https://lexica.art/}
\end{itemize}

The following models were used for object detection:
\begin{itemize}
    \item \href{https://huggingface.co/hustvl/yolos-base}{Huggingface YOLOS-base}
    \item \href{https://huggingface.co/ultralyticsplus/yolov8s}{Huggingface YOLO-v8}
\end{itemize}

\section{Detailed local SV example}
In this section, we are going to provide a detailed analysis of the local properties of CSED per frame for SV regarding the following sequence of Figure \ref{fig:examples}. This is an example of medium difficulty, according to the analysis followed in \cite{impartial}, as in the 4th frame the blue cylinder overlaps with the blue cube. We are going to compare the ground truth semantics of the sequence, corresponding to the ground truth frames (Figure \ref{fig:a}) with the generated semantics, corresponding to the generated frames (Figure \ref{fig:b}).
We, therefore, obtain the following results:
\paragraph{Frame k=1} \hfill \\
\textbf{\textit{Ground truth semantics}}: \{[small, brown, \textbf{metallic}, sphere]\} \\
\textbf{\textit{Generated semantics}}: \{[small, brown, \textbf{rubber}, sphere]\} \\
The two sequences differ by the highlighted semantic in the 3rd position: while the ground truth semantic is 'metallic', the generated is 'rubber', therefore $CSED_{k=1}$ proposes the replacement operation 'rubber' $\rightarrow$'metallic' with Edit cost = 2 =  $CSED_{k=1}$ in order for the generated sequence to become identical to the ground truth one. Moreover, as the transformation is an instance involving the \textit{Material} semantic, one more generation failure is added to the \textit{Material Loss} counter, which is going to provide global explanations regarding semantic synthesis failures for all test set frames.

For the 1st frame, Consistency Loss (CL) for the generated sequence is  $CL_{k=1}$ = 0, since there are $|\mathcal{C}|$=4 semantics in total (\textit{Material, Size, Shape, Color}), and 1 object containing $T=$4 semantics is placed in k=1 position in the sequence: $CL_{k=1} = p_{k=1} = |T|- |\mathcal{C}|\cdot k$=4-4=0.
\paragraph{Frame k=2} \hfill  \\
\textbf{\textit{Ground truth semantics}}: \{[small, brown, \textbf{metallic}, sphere], [small, brown, metallic, sphere]\} \\
\textbf{\textit{Generated semantics}}: \{[small, brown, \textbf{rubber}, sphere], [small, brown, metallic, sphere]\} \\
There is a difference in the semantic of the 3rd position, highlighted in bold: while the ground truth semantic is 'metallic', the generated is 'rubber', therefore $CSED_{k=2}$ proposes the replacement operation 'rubber' $\rightarrow$'metallic' with Edit cost = 2 = $CSED_{k=2}$. Moreover, as the transformation is an instance involving the \textit{Material} semantic, one more generation failure is added to the \textit{Material Loss} counter.

In the same time, CL will inevitably increase just by adding one more object containing $|\mathcal{C}|=$4 semantics. Therefore, the minimum increase of CL for CLEVR-SV when one object is added can be 4. Other than that, if there are more inconsistencies between $k=1$ and $k=2$ generated frames, $CL_{k=2}$ will increase. Therefore, we compare $k=1$ generated sequence $T = S_{k-1}$=\{[small, brown, rubber, sphere]\} with the $k=2$ generated sequence $S=S_k$=\{[small, brown, rubber, sphere], [\textbf{small, brown, metallic, sphere}]\}, where no extra differences are spotted. By applying equation \ref{eq:cl} for k=2 we obtain:

  $CL_{k=2} = p_{k=1} + D(S_{k=2}, T_{k=2}) = 0 + \textbf{I}\{small, brown, metallic, sphere\}$ = 0+4 = 4

\paragraph{Frame k=3} \hfill  \\
\textbf{\textit{Ground truth semantics}}: \{[small, brown, \textbf{metallic}, sphere], [small, brown, metallic, sphere], [large, blue, rubber, cube] \}\\
\textbf{\textit{Generated semantics}}: \{[small, brown, \textbf{rubber}, sphere], [small, brown, metallic, sphere], [large, blue, rubber, cube]\}\
The difference in the 3rd position semantic remains, therefore $CSED_{k=3}$ proposes the replacement operation 'rubber' $\rightarrow$'metallic' with Edit cost = 2 = $CSED_{k=3}$. Moreover, as the transformation is an instance involving the \textit{Material} semantic, one more generation failure is added to the \textit{Material Loss} counter. 

CL will take into account the comparison between $T=2$ generated sequence [small, brown, rubber, sphere, small, brown, metallic, sphere] and $k=3$ generated sequence \{[small, brown, rubber, sphere], [small, brown, metallic, sphere], [\textbf{large, blue, rubber, cube}]\}, which only differ by the addition of the \textit{large, blue, rubber, cube} in the third frame, thus yielding:

  $CL_{k=3} = p_{k=1} + D(S_{k=2}, T_{k=2})+ D(S_{k=3}, T_{k=3}) = 0 + 4+ \textbf{I}\{large, blue, rubber, cube\}$ = 0+4+4 = 8

\paragraph{Frame k=4} \hfill  \\
\textbf{\textit{Ground truth semantics}}: \{[small, brown, \textbf{metallic}, sphere], [small, brown, metallic, sphere], [large, blue, rubber, cube], [large, blue, metallic, \textbf{cylinder}]\} \\
\textbf{\textit{Generated semantics}}: \{[small, brown, \textbf{rubber}, sphere], [small, brown, metallic, sphere], [large, blue, rubber, cube], [large, blue, metallic, \textbf{sphere}]\}\\
Apart from the difference in the 3rd position semantic, for which $CSED_{k=4}$ proposes the replacement operation 'rubber' $\rightarrow$'metallic' with Edit cost = 2, there is also one difference in the last position semantic, indicating the transformation 'sphere' $\rightarrow$'cylinder' with Edit cost = 2. By aggregating the two transformations together, we obtain the total transformation for $k=4$: \{'rubber', 'sphere'\} $\rightarrow$ \{'metallic', 'cylinder'\} with Edit cost = 4 = $CSED_{k=4}$. Counters for \textit{Material Loss} and \textit{Shape Loss} will increase by 1 each.

For CL, the sequences corresponding to $k=3$ generated sequence \{[small, brown, rubber, sphere], [small, brown, metallic, sphere], [large, blue, rubber, cube]\} and $k=4$ generated sequence \{[small, brown, rubber, sphere], [small, brown, metallic, sphere], [large, blue, rubber, cube], [\textbf{large, blue, metallic, sphere}]\}, which only differ by the addition of the \textit{large, blue, metallic, sphere} item. Therefore $CL_{T=4}$= 4.

  $CL_{k=4} = p_{k=1} + D(S_{k=2}, T_{k=2})+ D(S_{k=3}, T_{k=3}) + D(S_{k=4}, T_{k=4}) = 0 + 4 + 4 + \textbf{I}\{large, blue, metallic, sphere\}$ = 0+4+4+4 = 12
By aggregating results, \textit{Story Loss (SL)} as a sum of per frame CSED costs will be:
\begin{align*}
SL = 2+2+2+4 = 10    
\end{align*}
and by averaging SL on all $L=4$ frames according to equation \ref{eq:avgsl}:
\begin{align*}
Average\ SL = \frac{1}{k}SL =  10/4 = 2.5
\end{align*}
For consistency, we follow equation \ref{eq:avgcl}:
\begin{align*}
Average\ CL = \frac{p_{k=1}}{k} + \frac{1}{k} \sum_{k=1}^{k=L}[CL_{k>1}\neq|\mathcal{C}| \cdot (k-1)] = 0+0=0
\end{align*}
The generated story of Figure \ref{fig:examples} is \textbf{fully consistent} as the \textit{Average CL} equals to 0, which is the ideal case. Therefore, no semantics are inserted, deleted, or altered within the generated sequence. 
It is however interesting that CL cannot capture the \textbf{faithfulness} error between the new item inserted in the 4th frame: while the ground truth item is a \textit{large, blue, metallic, cylinder}, the generated sequence inserts a \textit{large, blue, metallic, sphere}, but CL does not penalize more the difference in the semantic of the last position. On the contrary, SL is responsible to penalize for this error. Of course, the opposite scenario could be applicable in a different example, where CL would indicate an error that SL could not capture. This observation concludes that both metrics can be important, with SL focusing on faithfulness between ground truth and generated stories, while CL focuses on consistency between consequently generated frames. The better the model, the \textit{lower} both metrics should be on the global level.

\end{document}